\documentclass[journal]{IEEEtran}

\usepackage{bm}
\usepackage[cmex10]{amsmath}
\usepackage{amssymb}
\usepackage{amsthm}
\usepackage{graphicx}
\usepackage{multirow}
\usepackage[hyphens]{url}

\newtheorem{lemma}{Lemma}
\theoremstyle{remark}
\newtheorem{remark}{Remark}

\usepackage{algorithm}
\usepackage{algorithmicx}
\usepackage{algpseudocode}

    \algrenewcommand\algorithmicrequire{\textbf{Require:}}
    \algrenewcommand\algorithmicensure{\textbf{Postcondition:}}

\graphicspath{{./}{./figs/}}

\begin{document}

\title{Distributed Variational Bayesian Algorithms Over Sensor Networks
\thanks{Manuscript received ---. This work was supported by the National
Natural Science Foundation of China (Grant Nos. 61171153,
61571392, and 61471320) and the National Program for Special
Support of Eminent Professionals.}}

\author{Junhao~Hua, Chunguang~Li,~\IEEEmembership{Senior~Member,~IEEE}
\thanks{The authors are with the Department
of Information Science and Electronic Engineering, Zhejiang
University, Hangzhou 310027, China (C. Li is the corresponding
author, e-mail: cgli@zju.edu.cn).}
\thanks{Color versions of one or more of the figures in this paper are available
online at http://ieeexplore.ieee.org.}}

\markboth{IEEE TRANSACTIONS ON SIGNAL PROCESSING}%
{HUA AND LI: Distributed Variational Bayesian Algorithms Over
Sensor Networks}

\maketitle

\begin{abstract}
Distributed inference/estimation in Bayesian framework in the
context of sensor networks has recently received much attention
due to its broad applicability. The variational Bayesian (VB)
algorithm is a technique for approximating intractable integrals
arising in Bayesian inference. In this paper, we propose two novel
distributed VB algorithms for general Bayesian inference problem,
which can be applied to a very general class of
conjugate-exponential models. In the first approach, the global
natural parameters at each node are optimized using a stochastic
natural gradient that utilizes the Riemannian geometry of the
approximation space, followed by an information diffusion step for
cooperation with the neighbors. In the second method, a
constrained optimization formulation for distributed estimation is
established in natural parameter space and solved by alternating
direction method of multipliers (ADMM). An application of the
distributed inference/estimation of a Bayesian Gaussian mixture
model is then presented, to evaluate the effectiveness of the
proposed algorithms. Simulations on both synthetic and real
datasets demonstrate that the proposed algorithms have excellent
performance, which are almost as good as the corresponding
centralized VB algorithm relying on all data available in a
fusion center.
\end{abstract}


\IEEEpeerreviewmaketitle

\section{Introduction} \label{sec:1}

\IEEEPARstart{W}{ireless} sensor networks consist of an amount of
spatially distributed nodes/agents that have limited communication
capabilities due to energy and bandwidth constraints. Such
networks are well-suited to perform decentralized information
processing and inference tasks
\cite{taj2011distributed,liu2014distributed,shen2014distributed,chen2002source,lic2015freq}.
Distributed approach performs inference/estimation tasks locally
at each node using its local data and the information obtained
from its one-hop neighbors. Compared with the centralized
approach, it doesn't need a powerful fusion center. So it is more
flexible and provides robustness to node and/or link failures in a
network, in addition to saving communication resources and energy.
In view of this, many distributed inference/estimation algorithms
over networks have been proposed, such as consensus-based
\cite{olfati2004consensus, olfati2005consensus}, diffusion-based
\cite{cattivelli2010diffusion,Takahashi2010diffusion,liu2012diffusion,li2013diffusion}
and randomized gossip-based algorithms
\cite{boyd2006randomized,dimakis2010gossip}. With the development
of modern wireless sensor networks and the extension of their
application areas \cite{akyildiz2002survey, predd2006distributed},
the observations are getting larger and more complex. Therefore,
it is urgent to develop advanced in-network distributed algorithms
for data analysis.

The Bayesian modeling provides us with an elegant approach to
analyze massive data and its hidden structure. Several studies use
probabilistic graphical models for distributed Bayesian inference.
In \cite{paskin2005robust}, a distributed architecture using
message passing (or belief propagation) on a junction tree was
presented, which is constructed by the minimum spanning tree
algorithm. But exploring the junction tree itself is expensive in
low-cost networked systems. Combined with the belief propagation
(BP), an in-network variational message passing framework was
proposed for Markov random fields in \cite{dai2013structured}.
However, the convergence can not be guaranteed in loopy graphs and
it is intractable for complex models with non-Gaussian continuous
variables. For tractability, the nonparametric BP was developed
\cite{sudderth2010nonparametric,ihler2005nonparametric}, which
combines the ideas from Monte Carlo and particle filtering for
modeling uncertainty. However, the sampling-based technique is not
suitable for large and/or distributed datasets due to the heavy
computational cost.

The statistical inference tasks in the Bayesian framework often
suffer from the computational intractability of posterior beliefs
\cite{mackay1992practical,cooper1992bayesian}. To deal with it,
one of the most successful methods in practice is the variational
Bayesian (VB) approximation
\cite{vsmidl2006variational,attias1999inferring}.
In recent years, several scalable VB algorithms for
 massive and streaming data that arises in large scale
applications were developed \cite{wolfe2008fully,
hoffman2013stochastic,zhai2012mr}. However, none of
these algorithms are suitable for the networked systems. For
example, the authors in \cite{zhai2012mr} improve the scalability
of variational inference for latent Dirichlet allocation in the
MapReduce framework, but it needs a reducer (fusion center) and
multiple mappers (nodes). The goal of the present paper is to
design fully distributed variational Bayesian algorithms that can
perform almost as well as the centralized VB.

Several previous studies have developed distributed VB algorithms
for specific problems, especially for distributed density
estimation using the Gaussian mixture model. In
\cite{safarinejadian2010distributed}, an approach, which uses a
cyclic path to incrementally collect all local quantities
calculated at every node for the global estimation, was proposed.
This approach needs a prior knowledge of the network topology and
is not robust to node and/or link failures. In
\cite{mukherjee2008distributed,safarinejadian2011distributed}, the
distributed averaging strategies \cite{xiao2004fast} were adopted
to make a consensus among all nodes. In every VB step, each node
exchanges information and repeats the averaging iteration many
times, which exhausts the communication resources and energy.
What's more, these algorithms are not extensible because they need
to design local/global quantities carefully according to the
specific model.

In this paper, we develop two robust variational algorithms for
general Bayesian inference in a networked system. The first is
based on stochastic optimization and distributed averaging, called
as the distributed stochastic variational Bayesian algorithm
(dSVB). This approach is motivated by the earlier development on
the stochastic VB \cite{hoffman2013stochastic}, which can only
deal with the problem in the centralized scheme. We assume that
the data model belongs to conjugate exponential families and the
observed data is independent and identically distributed. Thus the
global lower bound (objective function) is decomposed by a set of
local lower bounds which are optimized in the natural parameter
space. The stochastic gradient approximation is then adopted for
local calculation, followed by a combination with its neighbors to
diffuse information over the entire network. Note that the natural
gradient \cite{amari1998natural} is adopted in our approach, since
the parameter space of a distribution is Riemannian rather than Euclidean.
The benefit of the stochastic gradient is
that it can gradually learn information from data through the VB procedure
with only one iteration in each VB step, which greatly reduces
the communication cost while holding high accuracy. Furthermore,
the quantity to be transmitted among neighbors is the natural
parameter vector, which provides a general form of the message.
Therefore, our algorithm is very general and can be automatically
derived.

The second novel distributed variational Bayesian algorithm is
based on the alternating direction method of multipliers
(dVB-ADMM). As a simple but powerful optimization algorithm, the
ADMM has been extended and developed for distributed convex
optimization in recent years
\cite{boyd2011distributed,forero2010consensus,forero2011distributed}.
It is a very robust algorithm and few assumptions are needed for
the convergence. To the best of our knowledge, the ADMM technique
has not been applied to the VB algorithm for distributed
inference/estimation. In this paper, the distributed VB algorithm
is derived by solving a constrained optimization problem.
The original variational objective function is equivalently
transformed into a simple convex function with respect to the
natural parameter vector. Importantly, the variational equality
constraints of distributions, which are hard to be measured in the
Riemannian space, are equivalently replaced by the equality
constraints of their natural parameter vector using the Euclidean
metric. A modified ADMM technique is then applied to solving this
optimization problem in the natural parameter space. The message
to be transmitted is also the natural parameter vector of a global
distribution, which has much lower dimension and smaller size than
the raw data.

In order to evaluate the effectiveness of
the proposed algorithms, examples on distributed clustering and
density estimation using Gaussian mixture model are presented.
Numerical simulations on both synthetic and real-world datasets
demonstrate that the proposed distributed approaches can perform
almost as well as the corresponding centralized one and outperform
the non-cooperation VB and non-stochastic-gradient based
distributed VB algorithm.
 Furthermore, the dVB-ADMM converges faster than the
dSVB.

The main contributions of this paper are summarized as follows.
\begin{itemize}
     \item We propose a general distributed VB framework for conjugate-exponential models over a network.
     \item We integrate the stochastic natural gradient with the alternating iterative procedure over a network
     and propose the distributed stochastic VB algorithm.
     \item We establish a constrained optimization formulation for distributed inference in natural parameter space
     and develop a fast distributed VB algorithm based on ADMM.
     \item We solve the distributed inference/estimation of a Bayesian Gaussian mixture model in WSNs
     based on the dSVB and dVB-ADMM, respectively.
\end{itemize}

The rest of the paper is organized as follows. Section \ref{sec:2}
states the problem and briefly reviews the traditional VB methods.
Section \ref{sec:3} presents the general distributed Bayesian
framework and then proposes the dSVB and the dVB-ADMM algorithms.
An application on the distributed inference/estimation of a
Bayesian Gaussian mixture model is then presented in Section
\ref{sec:4}. Section \ref{sec_exper_res} provides detailed
simulation results. Finally, conclusions are drawn in Section
\ref{sec_conclusions}.

\textit{Notation:}
In this paper, we use boldface letters for matrices (column vectors).
The superscript transposition $(\cdot)^T$ denotes transposition, and
$[\cdot]_{ij}([\cdot]_{i})$ denotes the $ij$-entry of a matrix ($i$-entry of a vector).
The operator $\mathbb{E}[\cdot]$ denotes expectation,  and
$|\cdot|$ denotes the determinant of a matrix or absolute value in case of a scalar.
Moreover, $\operatorname{tr}(\cdot)$ denotes the trace operator, $\nabla (\cdot)$ stands for the vector differential operator,
$\mathcal{N}(\cdot)$ is the Gaussian distribution, $\mbox{Dir}(\cdot)$ is the Dirichlet distribution,
$\mathcal{W}(\cdot)$ is the Wishart distribution, $\mathcal{NW}(\cdot)$ is normal-Wishart distribution,
and $\mbox{Mult}(\cdot)$ is the multinomial distributions.
Finally, $\Gamma(\cdot)$ is the Gamma function, and $\varphi(\cdot)$ is the Digamma function.
Other notations will be given if necessary.

\section{Problem Statement and Preliminaries} \label{sec:2}
We consider a sensor network consisting of $N$ agents distributed over a geographic region.
We use graphs to represent networks.
The considered undirected graph without a self-loop $\mathcal{G} = (\mathcal{V},
\mathcal{E})$ consists of a set of nodes $\mathcal{V} =
\{1,2,\dots,N\}$ and a set of edges $\mathcal{E}$, where each edge
$(i,j) \in \mathcal{E} $ connects an unordered pair of distinct
nodes. For each node $i \in \mathcal{V}$, let $ \mathcal{N}_i = \{
j | (i,j) \in \mathcal{E}\}$ be a set of neighbors of node $i$
(excluding node $i$ itself).

\begin{figure}[!t]
    \centering
    \includegraphics[width=2.0in]{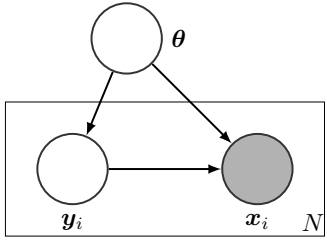}
    \caption{A graphical model with observations $\{\bm{x}_i\}$, local latent variables $\{\bm{y}_i\}$
        and global parameters $\bm{\theta}$. Circles represent random variables, Filled-in shapes indicate
        observed data and arrows describe conditional dependencies between variables.
        The $N$ represents the repetition of the variables in the plate.
        (Though not pictured, each variable may be a collection of multiple random variables.)}
    \label{fig_gm}
\end{figure}

Let us denote the observed dataset by $\bm{x}$. The data is
collected by the $N$ nodes of the network. Each node $i$ has $N_i$
$D$-dimension measurements $\bm{x}_i = \{\bm{x}_{ij},
j=1,2,\dots,N_i\}$, and the full observed data is made up of
measurements $ \bm{x}=\{\bm{x}_i,i=1,2,\dots,N\}$. Suppose the
observed data is  independent and identically distributed and
produced by a generative model whose form is given. Generally, a
generative model with unknown parameters often consists of a set
of latent variables. In particular, we use $\bm{y}$ to denote the
local latent variables and each node has a set of local latent
variables $\bm{y} = \{\bm{y}_1,\dots,\bm{y}_N\}$. For notational
convenience, we treat both the unknown parameters and the global
latent variables as the model parameters, denoted as
$\bm{\theta}$. We group model parameters and latent variables as
``unobserved variables'', denoted as $\bm{z} =
\{\bm{\theta},\bm{y}\}$. We assume that the $i$th observation
$\bm{x}_i$ and the $i$th local variable $\bm{y}_i$ are
conditionally independent, given parameters $\bm{\theta}$, of all
other observations and local latent variables. The graphical model
in \figurename{\ref{fig_gm}} captures the conditional dependencies
among local latent variables, parameters and observed variables.

Given the form of the generative model, VB is to approximate the
posterior of the unobserved variables, $P(\bm{z}|\bm{x})$, by a
more tractable distribution, $Q(\bm{z})$. It is found by
minimizing the Kullback-Leibler (KL) divergence between these two
distributions \cite{beal2003variational},
\setlength{\arraycolsep}{0.0em}
\begin{eqnarray} \label{KL-div}
 \mbox{KL}(Q(\bm{z}) || P(\bm{z} | \bm{x}))
  &{}={}& \int Q(\bm{z}) \log \frac{Q(\bm{z})}{P(\bm{z}| \bm{x})} d\bm{z}\nonumber\\
  &{}={}& -\mathbb{E}_{Q}[\log \frac{P(\bm{z}, \bm{x})}{Q(\bm{z})}] + \log P(\bm{x}) \nonumber \\
  &{}={}& -\mathcal{L}(Q(\bm{z})) + \log P(\bm{x}),
\end{eqnarray}
\setlength{\arraycolsep}{5pt}where the lower bound
$\mathcal{L}(Q)$ for the log evidence, $\log P(\bm{x})$, can be
rewritten as a summation of an energy term and an entropy term
(variational free energy),
\begin{equation} \label{lower_bound}
\mathcal{L}(Q(\bm{z})) = \mathbb{E}_{Q}[\log P(\bm{z},\bm{x})] + \mathbb{H}[Q(\bm{z})].
\end{equation}
Minimizing the KL divergence is equivalent to maximizing the
variational free energy since the log evidence is fixed with
respect to the variational distribution $Q$. Thus, the inference
task is presented as an optimization problem. In order to make
it tractable, this problem is then ``relaxed''.

The first relaxation is to use the naive mean field theory
\cite{saul1996mean}, which limits the optimization to be optimized
in a subset of distributions that are relatively easy to
characterize. Specifically, it assumes the variational posterior
of the unobserved variables can be factorized over some partitions
$\bm{z} = \{\bm{z}_1,\dots,\bm{z}_M\}$,
\begin{equation} \label{mean_field}
Q(\bm{z}) = \prod_{m=1}^M q_m(\bm{z}_m).
\end{equation}
Each partition $\bm{z}_m$ has its own variational distribution
$q_m(\bm{z}_m)$. In other words, each partition of the unobserved
variables is mutually independent given the data. With the mean
field assumption, the lower bound can be further decomposed into a
suitable form, which is given in the following lemma.
\begin{lemma}\label{lem_decomp_bound}
    For each probability distribution, $q_m$, the variational free energy (\ref{lower_bound})
    can be written as
    \begin{equation} \label{lower_bound_kl_form}
    \mathcal{L}(q_1,\dots,q_M) = - \mbox{KL} (q_m \|q_m^* ) + \mathbb{H}[q_{-m}] + \ln
    C.
    \end{equation}
    In (\ref{lower_bound_kl_form}),
    \begin{equation} \label{opt_star}
    q_m^* (\bm{z}_m) \triangleq \frac{1}{C} \exp \mathbb{E}_{q_{-m}}[ \ln P(\bm{z},\bm{x})],
    \end{equation}
    where $ C$ is a normalizing constant, and $q_{-m} (\bm{z}_{-m})$
    is the joint distribution of $\bm{z}_{-m}$ ($\bm{z}_{-m} \triangleq \bm{z} \setminus \bm{z}_m $
    denotes all unobserved variables except $ \bm{z}_m$).
\end{lemma}
Lemma \ref{lem_decomp_bound} is a general result whose proof can
be found in many literatures such as \cite{beal2003variational}.
Using the calculus of variations, it can be easily shown that the
``best'' distribution for $q_m$ is $q_m^*$, which depends on the
other distributions $q_{-m}$. The VB alternately updates each
variational posterior using
\begin{equation} \label{vb_update}
q_m = q_m^*, \forall m=1,2,\dots,M,
\end{equation}
with the other fixed. However, the VB update equation
(\ref{opt_star}) can be intractable since the expectation is in
fact an integral which usually has no analytical solution.
Usually, the optimization is further relaxed by approximating
the variational distribution to be optimized in the conjugate
exponential families of distributions, a broad class of
distributions that have been extensively studied in the statistics
literature \cite{brown1986fundamentals,wainwright2008graphical}.
Specifically, we assume the distributions of variables conditioned
on their parents, as presented in \figurename{\ref{fig_gm}}, are
drawn from the exponential family and are conjugate with respect
to the distributions over these parents variables.

A density function in the exponential families can be written in
the canonical form
\begin{equation} \label{exp_fam}
Q(\bm{z}) = h(\bm{z})\exp \{ \bm{\phi}^T \bm{u}(\bm{z})-A(\bm{\phi})\},
\end{equation}
where $\bm{u}(\bm{z})$ is a collection of functions of $\bm{z}$,
known as natural sufficient statistics, $\bm{\phi}$ is an
associated (natural) vector of canonical parameters, and
$A(\bm{\phi})$ acts as a normalization function ensuring
that the distribution integrates to unity for any given setting of
the parameters. The natural parameter vector $\bm{\phi}$ belongs to the
set (natural parameter space)
\begin{equation} \label{natural_param_sets}
 \Omega := \{ \bm{\phi}|A(\bm{\phi}) < +\infty \},
\end{equation}
where the log partition function $A(\cdot)$ is a convex function of $\bm{\phi}$, and
the domain $\Omega$ is a convex set \cite{wainwright2008graphical}.
We restrict our attention to regular exponential
families for which the domain $\Omega$ is an open set.

\begin{lemma} \label{lem_opt_exp_fam}
    In the conjugate-exponential families, the equation (\ref{opt_star})
    can be reparameterized as
    \begin{equation} \label{exp_fam_star}
    q_m^*(\bm{z}_m)  = h(\bm{z}_m)\exp \{ \bm{\phi}_m^{*T} \bm{u}(\bm{z}_m)-A(\bm{\phi}_m^*)\},
    \end{equation}
    where the natural parameter vector $\bm{\phi}_m^*$ is a function of
    expectations of related natural sufficient statistics.
\end{lemma}

Lemma \ref{lem_opt_exp_fam} is also a general result. Use the
properties of conjugate-exponential families, the variational
message passing (VMP) algorithm is formed to apply variational
inference into a Bayesian network. Following the VMP framework,
Lemma \ref{lem_opt_exp_fam} can be easily derived
\cite{winn2005variational}.

\begin{remark}
The expectation and covariance of the natural sufficient statistics
    $\bm{u}(\bm{z}_m)$ can be derived by
    \begin{subequations}
        \begin{align}
        &\mathbb{E}[\bm{u}(\bm{z})]= \nabla_{\bm{\phi}} A(\bm{\phi})  ,  \label{first_order_stat}\\
        &\mathbb{E}\left[ (\bm{u}(\bm{z}) - \mathbb{E}[\bm{u}(\bm{z})])
        (\bm{u}(\bm{z}) - \mathbb{E}[\bm{u}(\bm{z})])^T
        \right]=\nabla^2_{\bm{\phi}} A(\bm{\phi}), \label{sec_order_stat}
        \end{align}
    \end{subequations}
    which guarantee the computational tractability of
    the calculation of the natural parameter vector  $\bm{\phi}_m^*$ in Lemma \ref{lem_opt_exp_fam}.
\end{remark}

Since the form of variational distributions is known in prior and
stays unchanged in the iterations, the ``best'' distribution given
in (\ref{exp_fam_star}) can be totally determined and represented
by its natural parameter vector $\phi_m^*$. Therefore, the
variational update (\ref{vb_update}) can be simply written as
\begin{equation}\label{natural_param_1}
 \bm{\phi}_m = \bm{\phi}_m^*, \forall m=1,\dots,M.
\end{equation}
It tells us that the variational problem
(\ref{lower_bound_kl_form}) can be optimized directly through its
natural parameters in the parameter space rather than the
variational distribution in the probability space. We'll take
advantage of this fact in developing distributed VB methods below.

We can write (\ref{natural_param_1}) in a more familiar way.
The variational Bayesian algorithm alternates between maximizing
the lower bound with respect to the variational distribution of
latent variables $\bm{y}$ and that of the model parameters
$\bm{\theta}$. Therefore, the VB procedure (\ref{natural_param_1}) is like that of the
expectation-maximization (EM) algorithm \cite{dempster1977maximum}, consisting of two
iterative steps
\begin{subequations} \label{vbem_1}
\begin{align}
\text{VBE:\ \ } & \bm{\phi}_{y}^*   =\arg \max_{\bm{\phi}_{y}}
\mathcal{L}(\bm{\phi}_{y},\bm{\phi}_{\theta}^*), \label{vbe_1} \\
\text{VBM:\ \ } & \bm{\phi}_{\theta}^* = \arg \max_{\bm{\phi}_{\theta}}
\mathcal{L}(\bm{\phi}_{y}^*,\bm{\phi}_{\theta}) \label{vbm_1}.
\end{align}
\end{subequations}


In the next section, we propose two novel algorithms to solve the
optimization problems in the VB procedure (\ref{vbem_1}) in a
distributed fashion.

\section{Distributed Variational Bayesian Algorithms} \label{sec:3}

As mentioned in the Introduction, the total observed data $\bm{x}$
is not fully accessible in a single node, which makes the
distributed inference problem difficult. Fortunately, the joint
distribution $P(\bm{z},\bm{x})$ in the objective function
(\ref{lower_bound}) can be decomposed under the conditional
independence assumptions. It can be written as the product of the
conditional distributions, in which the likelihood of the observed
data at each node is separated, as shown below
\setlength{\arraycolsep}{0.0em}
\begin{eqnarray}\label{decomp_joint}
P(\bm{z},\bm{x}) &{}={}& P(\{\bm{y}_i\},\bm{\theta},\bm{x})
= P(\bm{\theta})\prod_i^N P(\bm{x}_i|\bm{y}_i,\bm{\theta})P(\bm{y}_i|\bm{\theta}) \nonumber \\
&{}={}& \Big(P(\bm{\theta})\prod_i^N P(\bm{y}_i|\bm{\theta}) \Big)\prod_i^N P(\bm{x}_i|\bm{y}_i,\bm{\theta}) \nonumber\\
&{}={}& P(\bm{z}) \prod_i^N P(\bm{x}_i|\bm{z}),
\end{eqnarray}
\setlength{\arraycolsep}{5pt}where $P(\bm{x}_i |\bm{y}_i, \bm{\theta})$
is replaced by $P(\bm{x}_i |\bm{z})$ in the third equation, for notational convenience.
Therefore, with a multiplication and division step,
the global lower bound $\mathcal{L}(Q)$ is replaced by an
average of the local lower bounds,
\setlength{\arraycolsep}{0.0em}
\begin{eqnarray}\label{decomp_lbound}
\mathcal{L}(Q(\bm{z})) &{}={}& \mathbb{E}_{Q}[\log P(\bm{z},\bm{x})] + \mathbb{H}[Q(\bm{z})] \nonumber\\
&{}={}& \frac{1}{N} \sum_{i=1}^N N \mathbb{E}_Q [ \log P(\bm{x}_i|\bm{z})] + \mathbb{E}_Q[\frac{P(\bm{z})}{Q(\bm{z})}] \nonumber \\
&{}={}& \frac{1}{N} \sum_{i=1}^N \mathcal{L}_i(Q(\bm{z})),
\end{eqnarray}
\setlength{\arraycolsep}{5pt} where
\begin{equation}
\mathcal{L}_i(Q)\triangleq  \mathbb{E}_Q [\log \frac{ P(\bm{x}_i|\bm{z})^N P(\bm{z})}{Q(\bm{z})}].
\end{equation}
The quantity $\mathcal{L}_i(Q)$ is the lower bound for the log
evidence of the observed data $\{\bm{x}_i\}_{N}$ at node $i$,
where $\{\cdot\}_N$ means that the observed data $\bm{x}_i$ is
replicated $N$  times. Namely,
\begin{equation}
\mathcal{L}_i(Q) \leq \log E_{P(\bm{z})} [P(\bm{x}_i|\bm{z})^N] = \log P(\{\bm{x}_i\}_N). \nonumber
\end{equation}
The first inequality is derived from Jensen's inequality
considering the concavity of the logarithmic function. The
equality holds if and only if $Q(\bm{z}) \sim P(\bm{x}_i|\bm{z})^N
P(\bm{z})$. Namely, $Q(\bm{z})$ is exactly the posterior of the
unobserved data $\bm{z}$ given the replicated observed data
$\{\bm{x}_i\}_N$. Although the lower bound can be decomposed, we
can not independently maximize the local lower bound at each node
to reach a global optimum. A key observation is that the global
lower bound for the log evidence of the full observed data is
definitely less than or equal to the averaged lower bound for the
log evidence of the local replicated observed data over all nodes.
Mathematically, \setlength{\arraycolsep}{0.0em}
\begin{eqnarray}
\max_{Q} \mathcal{L}(Q) &{}={}& \frac{1}{N} \sum_{i=1}^N \mathcal{L}_i(Q^*) \nonumber \\
&{ }\leq{ }& \frac{1}{N} \sum_{i=1}^N \mathcal{L}_i(Q_i^*)  \nonumber\\
&{}={}&  \frac{1}{N} \sum_{i=1}^N
\max_{Q}\mathcal{L}_i(Q),\label{global_less_local_bound}
\end{eqnarray}
\setlength{\arraycolsep}{5pt}where $Q^*$ is the optimal
variational distribution maximizing the global lower bound
$\mathcal{L}(Q)$, and $Q^*_i$ is the one maximizing the local
lower bound for the log evidence $\log P(\{\bm{x}_i\}_N)$. The
equality in the second line holds if and only if  $Q^*$ is also
the optimal solution for all the local lower bounds, which is not
always the case. So we can not find the optimal solution by
individually maximizing the local lower bound at each node, as
done in the third line, and a distributed approach has to be
designed to solve this problem.

As mentioned in the previous section, the lower bound can be
directly optimized in the space of natural parameters. The latent
variables are local variables, and we denote the natural parameter
vectors of latent variables at each node by
$\{\bm{\phi}_{y_i},i=1,\dots,N\}$. By replacing the lower bound
with the decomposed version (\ref{decomp_lbound}), the VB
procedure is rewritten as
\begin{subequations} \label{vb_procedure}
    \begin{align}
    \text{VBE:\ \ } & \bm{\phi}_{y_i}^*  = \arg \max_{\bm{\phi}_{y_i}}
    \mathcal{L}_i(\bm{\phi}_{y_i},\bm{\phi}_{\theta}^*), \forall\,\, i=1,\dots,N,  \label{vbe_2}\\
    \text{VBM:\ \ } & \bm{\phi}_{\theta}^* = \arg \max_{\bm{\phi}_{\theta}}
    \sum_{i=1}^N \mathcal{L}_i(\bm{\phi}_{y_i}^*,\bm{\phi}_{\theta}). \label{vbm_2}
    \end{align}
\end{subequations}
Given the global natural parameters $\bm{\phi}_{\theta}^*$, the
VBE step (\ref{vbe_2}) can be solved individually at each node.
The solution can be directly found using Lemma
\ref{lem_decomp_bound} and Lemma \ref{lem_opt_exp_fam} with a
slight modification that replaces $\bm{x}$ with $\{\bm{x}_i\}_N$.
However, the VBM step can not be solved straightforwardly. In the
next two subsections, two approaches are presented for
distributedly computing (\ref{vbm_2}).

\subsection{Distributed Stochastic VB} \label{subsec_dSVB}
Let's define a set of local natural parameter vectors
$\{\bm{\phi}_{\theta,i},i=1,\dots,N\}$ for each node. According to
Lemma \ref{lem_decomp_bound} and Lemma \ref{lem_opt_exp_fam}, the
``best'' variational distribution for the global model parameters
at each node with that of the latent variables fixed is
$q_{\theta,i}^{*}$,  and its natural parameter vector is
$\bm{\phi}_{\theta,i}^{*}$. In other words,  the local lower bound
$\mathcal{L}_i$ is maximized at $\bm{\phi}_{\theta,i}^{*}$ with
$\bm{\phi}_{y_i}^*$ fixed, namely
\begin{equation}\label{loc_opt}
\bm{\phi}_{\theta,i}^{*} = \arg \max_{\bm{\phi}_{\theta}}
\mathcal{L}_i(\bm{\phi}_{y_i}^{*},\bm{\phi}_{\theta}).
\end{equation}

Then the solution for (\ref{vbm_2}) is found by taking the
derivative of $\mathcal{L}_i$ with respect to $\bm{\phi}_{\theta}$,
\setlength{\arraycolsep}{0.0em}
\begin{eqnarray}
&& \sum_{i=1}^{N} \frac{\partial}{\partial  \bm{\phi}_{\theta}}
\mathcal{L}_i(\bm{\phi}_{y_i}^*,\bm{\phi}_{\theta})
 = -\sum_{i=1}^{N} \frac{\partial}{\partial  \bm{\phi}_{\theta}}
\mbox{KL} (q_{\theta} \|q_{\theta,i}^{*} ) \nonumber \\
&&  = -\sum_{i=1}^{N} \frac{\partial}{\partial \bm{\phi}_{\theta}}
\left( (\bm{\phi}_{\theta} -  \bm{\phi}_{\theta,i} ^{*})^T
\mathbb{E}_{\theta} [ \bm{u}(\theta)]-A(\bm{\phi}_{\theta})+
A(\bm{\phi}_{\theta,i}^{*})  \right) \nonumber\\
&& = -\sum_{i=1}^{N} \frac{\partial}{\partial  \bm{\phi}_{\theta}}
\left( ( \bm{\phi}_{\theta} - \bm{\phi}_{\theta,i} ^{*})^T
\nabla_{\bm{\phi}_{\theta}} A(\bm{\phi}_{\theta})
-A(\bm{\phi}_{\theta})  \right) \nonumber\\
&&  = -\sum_{i=1}^{N} \nabla_{\bm{\phi}_{\theta}}^2 A(\bm{\phi}_{\theta})
(\bm{\phi}_{\theta} -  \bm{\phi}_{\theta,i} ^{*}), \label{gradient_VBM}
\end{eqnarray}
\setlength{\arraycolsep}{5pt}where the first, second and third
equalities are derived from (\ref{lower_bound_kl_form}),
(\ref{exp_fam_star}) and (\ref{first_order_stat}), respectively.
Set the partial derivative to zero, we obtain the solution for the
VBM step (\ref{vbm_2})
\begin{equation} \label{vbm_sol}
\bm{\phi}_{\theta}^* = \frac{1}{N} \sum_{i=1}^{N} \bm{\phi}_{\theta,i}^{*}.
\end{equation}
It is an average of all the local optimal natural parameters
calculated at each node. If there exists a fusion center, which
can receive all local optimums from all nodes, then a centralized
VB is obtained. If a cyclic path through all the nodes could be
found, then an incremental VB algorithm can be derived. However,
neither can be applied in a low-cost networked system, since the
communication resources are limited and exploring the network
topology is hard and expensive.

Distributed averaging consensus approach has been proposed to
solve this kind of problem \cite{xiao2004fast,gu2008distributed}.
However, a key weakness of this approach is that too many
iterations are needed to reach a consensus in each VB step. A
diffusion-based EM algorithm is proposed in
\cite{weng2011diffusion} for distributed estimation of Gaussian
mixtures, in which each node diffuses its local statistics with
its neighbors only once per EM step (one-step averaging). One
might think that we can borrow this simple one-step averaging idea
to develop a distributed VB, and approximate (\ref{vbm_sol}) using
$ \bm{\phi}_{\theta,i}^{t} = \frac{1}{|\mathcal{N}_i| + 1}\sum_{j
\in \mathcal{N}_i \cup \{i\}} \bm{\phi}_{\theta,j}^{*,t}, \forall
i=1,\dots,N, $ where $t$ is a time instant and $|\mathcal{N}_i|$
is the degree of node $i$ (i.e., the number of neighbors of the
node $i$). However, it can not provide a good result when the
local data is imbalanced. Since the best value of
$\bm{\phi}_{\theta,i}^t$ in this procedure is totally determined
by its own and its neighbors' local latent variables
$\{\bm{\phi}_{y_j}^t\}_{j \in \mathcal{N}_i \cup \{i\}}$, which is
not a representative of the whole set of latent variables
$\{\bm{\phi}_{y_i}^t\}_{i=1}^N$ when the local data is imbalanced.

We propose a gradient-based method for distributed estimation.
Instead of using the first-order condition in the VBM step to
yield a local optimum $\bm{\phi}_{\theta,i}^{*}$ at each node, we
use a stochastic gradient method
\cite{hoffman2013stochastic,robbins1951stochastic}, followed by a
diffusion procedure \cite{cattivelli2010diffusion} to get the
global natural parameters (\ref{vbm_sol}) gradually and
approximately. We denote an intermediate quantity estimated by a
gradient ascent step at node $i$ as $\bm{\varphi}_{\theta,i}$.
After computing this intermediate quantity, a simple combination
step is followed. Namely, for each time $t$, the update equations
at node $i$ are
\begin{subequations} \label{grad_comb_1}
    \begin{align}
        \bm{\varphi}_{\theta,i}^{t} &=  \bm{\phi}_{\theta,i}^{t-1}
        + \eta_t \tilde{\nabla}_{\bm{\phi}_{\theta}}
        \mathcal{L}_i(\bm{\phi}_{y_i}^*,\bm{\phi}_{\theta,i}^{t-1}),
        \label{gradient_ascent_intermediate}\\
        \bm{\phi}_{\theta,i}^{t} &= \sum_{j\in \mathcal{N}_i \cup \{i\}} w_{ij}
        \bm{\varphi}_{\theta,j}^{t}, \label{combine_step_1}
    \end{align}
\end{subequations}
where $\tilde{\nabla}_{\bm{\phi}_{\theta}} \mathcal{L}_i$ denotes
the natural gradient \cite{amari1998natural} in a Riemannian space
(explained in following paragraphs), $\eta_t$ is the step size
satisfying \cite{robbins1951stochastic}
\begin{equation}\label{stepsize_condtion}
\sum \eta_t = \infty; {\ \ }  \sum \eta_t^2 < \infty,
\end{equation}
and  $\{w_{ij}\}$ are non-negative weights satisfying
\begin{equation}
 \sum_{j=1}^N w_{ij} = 1, w_{ij} = 0 \  \text{if} \ j \notin \mathcal{N}_i \cup \{i\}.
\end{equation}
There are many possible rules for choosing the weights
$\{w_{ij}\}$, such as the Metropolis, the Laplacian and the
nearest neighbors rules
\cite{olfati2004consensus,xiao2004fast,jadbabaie2003coordination}.
In addition, we can also use some strategies to optimize the
combination weights \cite{Takahashi2010diffusion,carli2008distributed}.

This approach is motivated by the diffusion LMS algorithm
\cite{cattivelli2010diffusion, lopes2007incremental}, which
addresses a distributed linear estimation problem in a cooperative
fashion. Though the steady-state performance of the diffusion
cooperation scheme has been well studied for the linear estimation
problem \cite{cattivelli2010diffusion}, theoretical performance
analysis of the proposed algorithm (\ref{grad_comb_1}) is hard
since the local lower bound $\mathcal{L}_i$ is more complex.
Instead of providing theoretical analysis, we show that the
procedure (\ref{grad_comb_1}) can be interpreted as a distributed
implementation of the stochastic variational inference
\cite{hoffman2013stochastic}. Each node runs a gradient ascent
step (\ref{gradient_ascent_intermediate}) using only the local
data, which is similar to the stochastic approximation based on
the subsample \cite{hoffman2013stochastic}. The combination step
(\ref{combine_step_1}), which diffuses all local estimates over
the entire network, can be considered as a procedure gradually
collecting global (all local) sufficient statistics (since
$\bm{\phi}_{\theta,i}$ is a function of expectations of related
natural sufficient statistics as shown in Lemma
\ref{lem_opt_exp_fam}) with the iterations of the VB procedure
(\ref{vb_procedure}). Compared with the one-step averaging
approach, our gradient-based approach takes the previous estimate
$\bm{\phi}_{\theta,i}^{t-1}$ into account in
(\ref{gradient_ascent_intermediate}), which is a result obtained
by diffusing all estimates among nodes over the entire network.
Therefore, the convergence point of the gradient ascent procedure
(\ref{gradient_ascent_intermediate}) is not a solution of the
local objective function $\mathcal{L}_i$ but that of the global
objective function (\ref{vbm_2}).

This approach is based on the natural gradient
\cite{amari1998natural} rather than the standard gradient. The
natural gradient of a function accounts for the information
geometry \cite{amari1985differential} of its parameter space,
using a Riemannian metric to adjust the direction of the standard
gradient. In variational inference, the natural gradients have
been used for nonlinear state space models
\cite{honkela2008natural}, Bayesian mixtures \cite{sato2001online}
and latent Dirichlet allocation \cite{hoffman2013stochastic}. The
Riemannian metric rather than the Euclidean metric is used in this
paper, since the latter can not properly scale the gradient under
the manifold of a probability distribution. As we know, different
parameters of a distribution have different roles, such as
location, shape and scale, and the effect of one parameter can be
mutually influenced by the other parameters. For example
\cite{hoffman2013stochastic}, the distributions
$\mathcal{N}(0,10^5)$ and $\mathcal{N}(10,10^5)$ are almost
indistinguishable, and the Euclidean distance between means of
those two distributions is $10$. In contrast, the distributions
$\mathcal{N}(0,0.01)$ and $\mathcal{N}(0.1,0.01)$ barely overlap,
but this is not reflected in the Euclidean distance between their
mean parameters, which is only $0.1$. It is showed in
\cite{amari1998natural} that the parameter space of a distribution
has a Riemannian metric structure. In this case, the natural
gradient can give the steepest direction.

In a Riemannian space of parameters,
the steepest ascent direction of the objective function
$\mathcal{L}_i(\bm{\phi}_{y_i}^*,\bm{\phi}_{\theta})$ for
$\bm{\phi}_{\theta}$ with $\bm{\phi}_{y_i}^*$ fixed
is given by
\begin{equation}
\tilde{\nabla}_{\bm{\phi}_{\theta}} \mathcal{L}_i(\bm{\phi}_{y_i}^*,\bm{\phi}_{\theta})
= G^{-1} (\bm{\phi}_{\theta}) \nabla_{\bm{\phi}_{\theta}} \mathcal{L}_i(\bm{\phi}_{y_i}^*,\bm{\phi}_{\theta}),
\end{equation}
where $G^{-1}$ is the inverse of the Riemannian metric
and $\nabla_{\bm{\phi}_{\theta}} \mathcal{L}_i$ is the standard gradient, which is given by
\begin{equation}
\nabla_{\bm{\phi}_{\theta}} \mathcal{L}_i = \nabla_{\bm{\phi}_{\theta}}^2 A(\bm{\phi}_{\theta})
(\bm{\phi}_{\theta} -  \bm{\phi}_{\theta,i} ^{*}).
\end{equation}
The metric $G(\bm{\phi}_{\theta})$ is the Fisher information
matrix of a distribution \cite{amari1982differential}.
Since $\bm{\phi}_{\theta}$ is a natural parameter of an
exponential family distribution $ P(\bm{\theta}|\bm{\phi}_{\theta})$,
the Fisher metric $G(\bm{\phi}_{\theta})$ defined by $ P(\bm{\theta}|\bm{\phi}_{\theta})$
is the second derivative of its log partition function
$A(\bm{\phi}_{\theta})$. Mathematically,
\setlength{\arraycolsep}{0.0em}
\begin{eqnarray}
G(\bm{\phi}_{\theta}) &{\ }={\ }&
\mathbb{E}\left[(\nabla_{\bm{\phi}_{\theta}} \log P(\bm{\theta}|\bm{\phi}_{\theta}))
(\nabla_{\bm{\phi}_{\theta}} \log P(\bm{\theta}|\bm{\phi}_{\theta}))^T\right] \nonumber\\
&{\ }={\ }& \mathbb{E}\left[ (\bm{u}(\bm{\theta})  - \mathbb{E}[\bm{u}(\bm{\theta})])
(\bm{u}(\bm{\theta})  - \mathbb{E}[\bm{u}(\bm{\theta})])^T\right] \nonumber\\
&{\ }={\ }&\nabla_{\bm{\phi}_{\theta}}^2 A(\bm{\phi}_{\theta}),\nonumber
\end{eqnarray}
\setlength{\arraycolsep}{5pt}where the third equality is derived
from (\ref{sec_order_stat}). The natural gradient of
$\mathcal{L}_i(\bm{\phi}_{y_i}^*,\bm{\phi}_{\theta})$ w.r.t.
$\bm{\phi}_{\theta}$ is then simplified as
\setlength{\arraycolsep}{0.0em}
\begin{eqnarray}
\tilde{\nabla}_{\bm{\phi}_{\theta}}\mathcal{L}_i(\bm{\phi}_{y_i}^*,\bm{\phi}_{\theta})  &{\ }={\ }&
\nabla_{\bm{\phi}_{\theta}}^{-2} A(\bm{\phi}_{\theta})
\nabla_{\bm{\phi}_{\theta}} \mathcal{L}_i(\bm{\phi}_{y_i}^*,\bm{\phi}_{\theta,i}^{t-1}) \nonumber\\
 &{}={}& \bm{\phi}_{\theta,i} ^{*} - \bm{\phi}_{\theta,i}^{t-1}. \label{natural_grad_local_obj}
\end{eqnarray}
\setlength{\arraycolsep}{5pt}Substituting it into
(\ref{gradient_ascent_intermediate}), we obtain the procedure for
computing the global natural parameters at each node $i$
\begin{subequations}\label{adapt_comb}
\begin{align}
\bm{\varphi}_{\theta,i}^{t} &{}=  \bm{\phi}_{\theta,i}^{t-1} \label{adapt_step}
+ \eta_t (\bm{\phi}_{\theta,i} ^{*,t} - \bm{\phi}_{\theta,i}^{t-1}), \\
\bm{\phi}_{\theta,i}^{t} &{}=  \sum_{j\in \mathcal{N}_i \cup \{i\}} w_{ij} \bm{\varphi}_{\theta,j}^{t}. \label{combine_step}
\end{align}
\end{subequations}

At time instant $t$, each node calculates its intermediate
quantity $\bm{\varphi}_{\theta,i}^{t}$ using its local data, then
transfers it to its neighbors and also receives messages
$\{\bm{\varphi}_{\theta,j}^{t},  j \in \mathcal{{N}}_i\}$ from the
neighbors. Since usually the size and dimension of the
intermediate quantities are much smaller than the raw data, it
saves the communication resources and energy to a great extent.
Rather than reaching a consensus in each VB step, the network
diffuses the information along with the VB iterations so that
(\ref{adapt_comb}) only needs to iterate once in a single VB step.
To further analyze this process, we rewrite (\ref{adapt_comb})
into a single update:
\begin{equation} \label{dSVB_single_update}
\bm{\phi}_{\theta,i}^{t}
=\sum_{j\in \mathcal{N}_i \cup \{i\}} w_{ij} \bm{\phi}_{\theta,j}^{t-1}
+ \eta_t \sum_{j\in \mathcal{N}_i \cup \{i\}} w_{ij}  (\bm{\phi}_{\theta,j} ^{*,t} - \bm{\phi}_{\theta,j}^{t-1}).
\end{equation}
The role of the first term in (\ref{dSVB_single_update}) is to
diffuse information over the entire network. The second term
gradually updates estimate using the local data and the
information received from its neighbors. The step size $\eta_t$ is
a trade-off between the diffusion speed (the first term) and the
learning speed (the second term). Although the residuals in the
second term would not strictly be eliminated, a sufficiently small
step size $\eta_t$ ensures a small steady state error. But a small
step size also decreases the convergence speed. Conversely, a
large step size improves the rate of convergence but might lead to
instability.
\begin{remark}[On the selection of the step size $\eta_t$]
    We suggest to use a time-varying step size for the natural gradient,
    \begin{equation} \label{dSVB_step_size}
    \eta_t = \frac{1}{d_0 +\tau t},\   1\leq d_0, 0 < \tau < 1,
    \end{equation}
    which satisfies the conditions (\ref{stepsize_condtion}). In (\ref{dSVB_step_size}),
    the forgetting rate $\tau$ controls the decreasing speed of the step size
    and the parameter $d_0$ down-weights early iterations \cite{hoffman2013stochastic}.
    In Section \ref{sec_exper_res}, we empirically fix  $d_0=1$. We explore a variety of forgetting
    rates, and numerical simulations show that $\tau \in [0.1, 0.3]$ could give a good performance
    for kinds of applications/problems.
\end{remark}

For clarity, the distributed stochastic variational Bayesian algorithm (dSVB)
is summarized in Algorithm \ref{alg_dSVB}.

\begin{algorithm}
    \caption{ The dSVB algorithm \label{alg_dSVB}}
    \begin{algorithmic}[1]
        \Require{Node $i$ observes data $\bm{x}_i$.
            The natural parameters are initialized using non-informative priors. }
        \Statex
        \State Set the tuning parameter $\tau$ appropriately.
        \For{$t \gets 1,2,\dots$}  \Comment{$t$: time step}
            \For{\textbf{all} $i=1,\dots,N$}
                \State $\bm{\phi}_{y_i}^{*,t}  = \arg \max_{\bm{\phi}_{y_i}}
                \mathcal{L}_i(\bm{\phi}_{y_i},\bm{\phi}_{\theta,i}^{t-1})$.\Comment{VBE}
                \State $\bm{\phi}_{\theta,i}^{*,t} = \arg \max_{\bm{\phi}_{\theta}}
                \mathcal{L}_i(\bm{\phi}_{y_i}^{*,t},\bm{\phi}_{\theta})$.
                \State Compute $\bm{\varphi}_{\theta,i}^{t}$ via (\ref{adapt_step}). \Comment{Natural gradient}
                \State Broadcast $\bm{\varphi}_{\theta,i}^{t}$ to all neighbors in $\mathcal{N}_i$.
            \EndFor
            \For{\textbf{all} $i=1,\dots,N$}
                \State Compute $\bm{\phi}_{\theta,i}^{t}$ via (\ref{combine_step}). \Comment{Combination}
            \EndFor
        \EndFor
    \end{algorithmic}
\end{algorithm}

\subsection{Distributed VB based on ADMM}
In this subsection, we reformulate the VBM step as a constrained
minimization problem, and then we use a well-studied optimization
technique, the alternating direction method of multipliers (ADMM)
\cite{boyd2011distributed}, to solve it. By adding the consensus
constraints that local variables $\{\bm{\phi}_{\theta,i}\}$ agree
upon with its neighbors, the maximization problem (\ref{vbm_2}) of
the VBM step can be formulated as, \setlength{\arraycolsep}{0.0em}
\begin{eqnarray}
\min_{\{\bm{\phi}_{\theta,i}\},\{\bm{\varphi}_{\theta,i,j}\}} \  && \
- \sum_{i=1}^N  \mathcal{L}_i(\bm{\phi}_{y_i}^*,\bm{\phi}_{\theta,i}) \label{vbm_admm} \\
 \text{s.t.} \ \ &{ }& \bm{\phi}_{\theta,i} = \bm{\varphi}_{\theta,i,j},  \ i=1,\dots,N,
 \ j \in \mathcal{N}_i \nonumber \\
 \text{s.t.} \ \ &{ }& \bm{\varphi}_{\theta,i,j} = \bm{\phi}_{\theta,j},  \ i=1,\dots,N,
 \ j \in \mathcal{N}_i, \nonumber
\end{eqnarray}
\setlength{\arraycolsep}{5pt}where the auxiliary variables
$\bm{\varphi}_{\theta,i,j} $ decouple local variables
$\bm{\phi}_{\theta,i}$ at node $i$ from those of their neighbors
$j\in \mathcal{N}_i$. With the assumption that the network remains
connected, the consensus constraints guarantee that problem
(\ref{vbm_2}) and (\ref{vbm_admm}) are equivalent. Since there
exists a path between any two nodes in the network, the consensus
constraints imply that the local variables in these two nodes are
equal. Since the pair of nodes is arbitrary, any feasible solution
for (\ref{vbm_admm}) is also a solution for (\ref{vbm_2}).

The ADMM technique can not be used for solving (\ref{vbm_admm}) in
a straightforward manner. The objective function in
(\ref{vbm_admm}) is a function of a variational distribution,
whose variational parameters have the Riemannian metric structure,
as discussed in the previous subsection. In contrast, the equality
constraints and quadratic penalty terms in the standard method of
multipliers are in fact using the Euclidean metric. One possible
way for solving this inconsistence is to replace the penalty term
with a more general deviation penalty, whose parameter space also
has the Riemannian character, such as the one derived from a
Bregman divergence \cite{banerjee2005clustering}. Unfortunately,
to the best of our knowledge, there is no proof of the convergence
of ADMM with nonquadratic penalty terms available currently.

Therefore, we turn our attention to designing an objective
function using the Euclidean metric that is equivalent to
(\ref{vbm_admm}). Using the results derived in
(\ref{natural_grad_local_obj}), it's easy to show that the natural
gradient of (\ref{vbm_admm}) with respect to
$\bm{\phi}_{\theta,i}$ is $\bm{\phi}_{\theta,i} -
\bm{\phi}_{\theta,i} ^{*}$. This implies that the following
relation holds for all $i$,
\begin{equation*}
\nabla_{\bm{\phi}_{\theta,i}}(\frac{1}{2}\| \bm{\phi}_{\theta,i} - \bm{\phi}_{\theta,i} ^{*}\|_F^2 )
= -\tilde{\nabla}_{\bm{\phi}_{\theta,i}} \mathcal{L}_i(\bm{\phi}_{y_i}^*,\bm{\phi}_{\theta,i}),
\end{equation*}
where $\|\cdot\|_F$ is the Frobenius norm. Consequently, the
original problem (\ref{vbm_admm}) can be recast as the following
formulation, \setlength{\arraycolsep}{0.0em}
\begin{eqnarray}  \label{admm_problem}
\min_{\bm{\phi}_{\theta,i},\bm{\varphi}_{\theta,i,j}}\  && \
\frac{1}{2} \sum_{i=1}^N \|\bm{\phi}_{\theta,i} - \bm{\phi}_{\theta,i} ^{*} \|_F^2 \label{vbm_admm_2} \\
\text{s.t.} \ \ &{ }& \bm{\phi}_{\theta,i} = \bm{\varphi}_{\theta,i,j}, i=1,\dots,N,\ j \in \mathcal{N}_i, \nonumber\\
\text{s.t.} \ \ &{ }& \bm{\varphi}_{\theta,i,j} = \bm{\phi}_{\theta,j}, i=1,\dots,N,\ j \in \mathcal{N}_i. \nonumber
\end{eqnarray}
\setlength{\arraycolsep}{5pt}

We could use a dual decomposition method for the separated
objective function (\ref{vbm_admm_2}). However, using the
augmented Lagrangian can bring robustness to the dual ascent
method \cite{boyd2011distributed}, so we use this method below.
Let $\bm{\lambda}_{ij1}$ and $\bm{\lambda}_{ij2}$ denote the
Lagrange multipliers corresponding to the constraints
$\bm{\phi}_{\theta,i} = \bm{\varphi}_{\theta,i,j}$ and
$\bm{\varphi}_{\theta,i,j} = \bm{\phi}_{\theta,j}$, the augmented
Lagrangian function is formulated as
\setlength{\arraycolsep}{0.0em}
\begin{eqnarray}
&&\mathcal{L}_{\rho}(\{\bm{\phi}_{\theta,i}\},\{\bm{\varphi}_{\theta,i,j}\},\{\bm{\lambda}_{ij1},\bm{\lambda}_{ij2}\})  \nonumber\\
&&= \frac{1}{2} \sum_{i=1}^N\| \bm{\phi}_{\theta,i} - \bm{\phi}_{\theta,i} ^{*}\|_F^2 \nonumber\\
&& {\ \ } + \sum_{i=1}^N\sum_{j \in \mathcal{N}_i}
\Big( \operatorname{tr}\big(\bm{\lambda}_{ij1}^T(\bm{\phi}_{\theta,i} - \bm{\varphi}_{\theta,i,j})\big) \nonumber\\
&& {\ \ } + \operatorname{tr}\big(\bm{\lambda}_{ij2}^T( \bm{\varphi}_{\theta,i,j} - \bm{\phi}_{\theta,j})\big)  \Big) \nonumber\\
&& {\ \ } + \frac{\rho}{2} \sum_{i=1}^N\sum_{j \in \mathcal{N}_i}
\Big( \|\bm{\phi}_{\theta,i} - \bm{\varphi}_{\theta,i,j}\|_F^2 + \|\bm{\varphi}_{\theta,i,j} - \bm{\phi}_{\theta,j})\|_F^2 \Big),  {\ \ \ \ } \label{eqn_lagrangian}
 \end{eqnarray}
\setlength{\arraycolsep}{5pt}where $\rho>0$ is a penalty
parameter. The ADMM solves (\ref{eqn_lagrangian}) in a cyclic
fashion by minimizing $\mathcal{L}_{\rho}$ with respect to the
local variables $\{\bm{\phi}_{\theta,i}\}$ and auxiliary variables
$\{\bm{\varphi}_{\theta,i,j}\}$, followed by a gradient ascent
step over the dual variables
$\{\bm{\lambda}_{ij1},\bm{\lambda}_{ij2}\}$. Taking the derivative of
the Lagrangian (\ref{eqn_lagrangian}) with respect to each
variable, and setting the derivative to zero, we get a closed-form
solution $\forall\, i=1,\dots,N, \forall\, j \in \mathcal{N}_i$,
\begin{subequations} \label{all_admm_1}
    \begin{align}
    \bm{\phi}_{\theta,i}^{t} &=
    \frac{\bm{\phi}_{\theta,i}^{*,t} + \sum_{j\in \mathcal{N}_i}
        \left(
        \bm{\lambda}_{ji2}^{t-1} - \bm{\lambda}_{ij1}^{t-1} +
        \rho( \bm{\varphi}_{\theta,i,j}^{t-1} + \bm{\varphi}_{\theta,j,i}^{t-1} )
        \right) }
    {1+2\rho N_i}, \label{eqn_admm_1} \\
    \bm{\varphi}_{\theta,i,j}^t
    &= \frac{1}{2 \rho} (\bm{\lambda}_{ij1}^{t-1}-\bm{\lambda}_{ij2}^{t-1})
    + \frac{1}{2} (\bm{\phi}_{\theta,i}^t + \bm{\phi}_{\theta,j}^t), \label{eqn_admm_2}\\
    \bm{\lambda}_{ij1}^{t} &= \bm{\lambda}_{ij1}^{t-1} + \rho (\bm{\phi}_{\theta,i}^t - \bm{\varphi}_{\theta,i,j}^t), \label{eqn_admm_3}\\
    \bm{\lambda}_{ij2}^{t} &= \bm{\lambda}_{ij2}^{t-1} + \rho (\bm{\varphi}_{\theta,i,j}^t - \bm{\phi}_{\theta,j}^t). \label{eqn_admm_4}
    \end{align}
\end{subequations}
Substituting (\ref{eqn_admm_2}) into (\ref{eqn_admm_3}) and
(\ref{eqn_admm_4}), we have
\begin{subequations}
    \begin{align}
    \bm{\lambda}_{ij1}^{t} &=\frac{1}{2}(\bm{\lambda}_{ij1}^{t-1}+\bm{\lambda}_{ij2}^{t-1})
    + \frac{\rho}{2}(\bm{\phi}_i^t - \bm{\phi}_j^t), \label{eqn_admm_5}\\
    \bm{\lambda}_{ij2}^{t} &= \frac{1}{2}(\bm{\lambda}_{ij1}^{t-1}+\bm{\lambda}_{ij2}^{t-1})
    + \frac{\rho}{2}(\bm{\phi}_i^t - \bm{\phi}_j^t). \label{eqn_admm_6}
    \end{align}
\end{subequations}
All the Lagrange multipliers are initialized to zeros at each
node. By mathematical induction, we know that
$\bm{\lambda}_{ij1}^{t} = \bm{\lambda}_{ij2}^{t} $ and
$\bm{\lambda}_{ij1}^{t} = -\bm{\lambda}_{ji1}^{t}, \forall
i=1,\dots,N, j \in \mathcal{N}_i$ for any time instant $t$.
Therefore, the auxiliary variable $\bm{\varphi}_{\theta,i,j}^t$
can be expressed as
\begin{equation} \label{auxiliary_update}
\bm{\varphi}_{\theta,i,j}^t   = \frac{1}{2} (\bm{\phi}_{\theta,i}^t + \bm{\phi}_{\theta,j}^t).
\end{equation}
Substituting it into (\ref{eqn_admm_1}), we have
\begin{equation}\label{admm_primal_update}
\bm{\phi}_{\theta,i}^{t} =
\frac{\bm{\phi}_{\theta,i}^{*,t} -2 \bm{\lambda}_{i}^{t-1}
    + \rho\sum_{j\in \mathcal{N}_i}
    ( \bm{\phi}_{\theta,i}^{t-1} + \bm{\phi}_{\theta,j}^{t-1} ) }
{1+2\rho N_i},
\end{equation}
where $\bm{\lambda}_{i}^{t} \triangleq \sum_{j \in \mathcal{N}_i}
\bm{\lambda}_{ij1}^{t}$ denotes a local aggregate Lagrange
multiplier. By substituting (\ref{auxiliary_update}) into
(\ref{eqn_admm_5}), we obtain an iteration equation for the local
aggregate Lagrange multiplier
\begin{equation} \label{admm_dual_update}
\bm{\lambda}_{i}^{t} = \bm{\lambda}_{i}^{t-1} +
\rho/2 \sum_{j \in \mathcal{N}_i} (\bm{\phi}_{\theta,i}^{t} -
\bm{\phi}_{\theta,j}^{t}).
\end{equation}
Thus, the auxiliary variables $\{\bm{\varphi}_{\theta,i,j}\}$ now
is eliminated. Alternating between (\ref{admm_primal_update}) and
(\ref{admm_dual_update}) for all nodes solves the optimization in
the VBM step. Thus, we solve the problem (\ref{vb_procedure}) in a
distributed fashion.

We restrict the ADMM procedure (\ref{all_admm_1}) to running only
one time in each VBM step in order to save communication resources
and energy. However, a numerical issue will arise, which will not
happen when the ADMM runs multiple times until it converges. Note
that unlike the procedure (\ref{adapt_step}), (\ref{combine_step})
in the dSVB approach, the update equation
(\ref{admm_primal_update}) is not a convex operation because the
sign in front of the multipliers $\bm{\lambda}_i^t$ is
negative. As a result, the updated natural parameters
$\bm{\phi}_{\theta,i}^t$ may not belong to the convex set $\Omega$
defined in (\ref{natural_param_sets}) though
$\bm{\phi}_{\theta,i}^{t-1}$ is in the convex set. In other words,
the log partition $A(\bm{\phi}_{\theta,i}^t)$ may become infinite.
For example, the covariance matrix in a normal distribution may
become negative definite.

To handle with this issue, a possible solution is to use the
projected gradient algorithm \cite{nesterov2004introductory}.
Using this method, the procedure of minimizing the augmented
Lagrangian function (\ref{eqn_lagrangian}) w.r.t.
$\bm{\phi}_{\theta,i}$ subject to the constraint
$\bm{\phi}_{\theta} \in \Omega$ can be written as two steps
\begin{subequations} \label{prmial_update_with_proj}
    \begin{align}
        \bm{\hat{\phi}}_{\theta,i}^{t} &=
        \frac{\bm{\phi}_{\theta,i}^{*,t} -2 \bm{\lambda}_{i}^{t-1}
            + \rho\sum_{j\in \mathcal{N}_i}
            ( \bm{\phi}_{\theta,i}^{t-1} + \bm{\phi}_{\theta,j}^{t-1} ) }
        {1+2\rho N_i}, \label{admm_primal_update_2}\\
        \bm{\phi}_{\theta,i}^{t} &= \arg \min_{\bm{\phi}_{\theta} \in \Omega}
        \|\bm{\phi}_{\theta} - \bm{\hat{\phi}}_{\theta,i}^{t}\|_F^2, \label{project_step}
    \end{align}
\end{subequations}
where the first step is exactly the same as
(\ref{admm_primal_update}), and the second step is the projection
of $\bm{\hat{\phi}}_{\theta,i}^{t}$ onto set $\Omega$.
Unfortunately, this extra projection step may make the optimization variable
$\bm{\phi}_{\theta,i}^t$ far away from the optimal value (\ref{vbm_sol})
when the point $\bm{\hat{\phi}}_{\theta,i}^{t}$ is far away from the domain $\Omega$.
Hence, we do not simply use this method.

Instead, we use a trick to handle with this numerical issue. We
introduce a time-varying parameter $\kappa_t$ to control the
evolution of the dual variables $\bm{\lambda}_i$. By replacing the
step size $\rho/2$  with $\kappa_t \rho/2$, the new gradient
ascent iteration for the multipliers becomes
\begin{equation}
    \bm{\lambda}_{i}^{t} = \bm{\lambda}_{i}^{t-1} +
    \kappa_t \rho/2 \sum_{j \in \mathcal{N}_i} (\bm{\phi}_{\theta,i}^{t} - \bm{\phi}_{\theta,j}^{t}).\label{admm_dual_update_2}
\end{equation}
Note that the local optimums $\{\bm{\phi}_{\theta,i} ^{*}\}$ among
nodes could be very different, and the residuals (i.e. dual
variables) among nodes could be very large at the very beginning
stage of the ADMM iterations. Thus the difference between
$\bm{\lambda}_{i}^{t}$ and $\bm{\lambda}_{i}^{t-1}$ could be very
large. The fact that the sign in front of
$\bm{\lambda}_{i}^{t}$ is negative in (\ref{admm_primal_update})
may make the point $\bm{\phi}_{\theta,i}^t$ not in the
convex set $\Omega$ after the step (\ref{admm_primal_update}). So
we set a small value for $\kappa_t$ at the very beginning stage to
ensure $\bm{\phi}_{\theta,i}^t$ being in the interior of the set
$\Omega$ (or being not in the interior but close to it, hence the
projection step (\ref{project_step}) can be applied), and
gradually increase it until it reaches $1$. We use the following
simple equation for updating the time-varying scalar factor,
\begin{equation} \label{control_factor}
\kappa_t = 1 - \frac{1}{(1+\xi t)^2},\ 0 < \xi < 1,
\end{equation}
where the parameter $\xi$ controls the increasing speed. Although
the optimal value of $\xi$ may depend on the observations, numerical
simulations show that a small value of $\xi$ is usually a good choice.
We set $\xi = 0.05$ in the following examples.

The distributed VB algorithm based on ADMM (dVB-ADMM) is
summarized in the Algorithm \ref{alg_dVB_ADMM}.

\begin{algorithm}
    \caption{ The dVB-ADMM algorithm \label{alg_dVB_ADMM}}
    \begin{algorithmic}[1]
        \Require{Node $i$ observes data $\bm{x}_i$.
            The natural parameters are initialized using non-informative priors. }
        \Statex
        \State Set the penalty parameter $\rho$ appropriately.
        \State Set $\bm{\lambda}_{i} = \bm{0}, \forall i$.
        \For{$t \gets 1,2,\dots$}  \Comment{$t$: time step}
        \For{\textbf{all} $i=1,\dots,N$}
        \State $\bm{\phi}_{y_i}^{*,t}  = \arg \max_{\bm{\phi}_{y_i}}
        \mathcal{L}_i(\bm{\phi}_{y_i},\bm{\phi}_{\theta,i}^{t-1})$.\Comment{VBE}
        \State $\bm{\phi}_{\theta,i}^{*,t} = \arg \max_{\bm{\phi}_{\theta}}
        \mathcal{L}_i(\bm{\phi}_{y_i}^{*,t},\bm{\phi}_{\theta})$.
        \State Compute $\bm{\phi}_{\theta,i}^{t}$ via (\ref{prmial_update_with_proj}). \Comment{Primal update}
        \State Broadcast $\bm{\phi}_{\theta,i}^{t}$ to all neighbors in $\mathcal{N}_i$.
        \EndFor
        \For{\textbf{all} $i=1,\dots,N$}
        \State Compute $\bm{\lambda}_{i}^{t}$ via (\ref{admm_dual_update_2}). \Comment{Dual update}
        \EndFor
        \EndFor
    \end{algorithmic}
\end{algorithm}

\begin{remark}[On the selection of the penalty parameter $\rho$] \label{remark_rho}
Unlike the step size $\eta_t$ in the dSVB algorithm, the step size
$\rho$ in (\ref{admm_dual_update}) is not necessarily to be
time-varying and to satisfy the condition
(\ref{stepsize_condtion}). So we can choose a fixed step size
$\rho$. Numerical simulations in Section \ref{sec_exper_res} will
illustrate that $\rho$ affects the convergence speed of the
algorithm. To explain this, we refer to $\bm{r}_i^t = \sum_{j \in
\mathcal{N}_i} (\bm{\phi}_i^t - \bm{\phi}_j^t)$ as the primal
residual and $\bm{s}_i^{t} = \rho \sum_{j \in \mathcal{N}_i
\cup i}(\bm{\phi}_j^t-\bm{\phi}_j^{t-1})$ as the dual residual at
iteration $t$, which can be derived from the primal and dual
feasibility conditions for problem (\ref{admm_problem}). A large
value of $\rho$ leads to a large penalty on violations of the
primal feasibility and so tends to produce a small primal residual. 
Conversely, a small value of $\rho$ would
produce a small dual residual, which in turn may induce a large
primal residual. Numerical simulations in Section
\ref{sec_exper_res} suggest that a relatively small value of
$\rho$ is preferable.
\end{remark}

\begin{remark}[On the convergence of the dVB-ADMM]
Since the distributions here belong to the conjugate-exponential
family, the space of natural parameters is always convex
\cite{wainwright2008graphical}. Thus, the ``best'' natural
parameter $\bm{\phi}_m^{*}$ in Lemma \ref{lem_opt_exp_fam} can
always be found. In other words, the solutions of the VBE step
(\ref{vbe_2}) and the local optimization problem (\ref{loc_opt})
at each node exist and can be easily obtained. In the dVB-ADMM,
the solution of (\ref{loc_opt}), in obtaining which we uses the
solution of (\ref{vbe_2}), is the local quantities
$\{\bm{\phi}_{\theta,i}^{*,t}\}$ in (\ref{admm_primal_update}).
The VBE step (\ref{vbe_2}) and the local optimizing step
(\ref{loc_opt}) can be viewed as a part of the alternating
procedure of the ADMM. As we know, the ADMM is proved to converge
in the context of distributed consensus problems
\cite{boyd2011distributed,erseghe2011fast,forero2011distributed}.
So, with appropriate penalty parameter, the ADMM iterations
(\ref{vbe_2}), (\ref{loc_opt}), (\ref{admm_primal_update}) and
(\ref{admm_dual_update_2}) can guarantee convergence.
\end{remark}

\section{Distributed VB for Gaussian Mixture Models} \label{sec:4}
In recent years, the mixture models over sensor networks have been
studied and applied to distributed density estimation, distributed
clustering, etc.
\cite{safarinejadian2010distributed,gu2008distributed,
weng2011diffusion}. A standard method for this
inference/estimation problem is based on the EM algorithm under
the maximum likelihood (ML) framework. However, ML is well-known
for its tendency toward overfitting the data and its preference of
complex models. A fully Bayesian treatment of mixture modelling
can avoid overfitting by integrating out the parameters and
identity the optimal structure of models by automatically
penalizing the complex model with a lower posterior probability.
Unfortunately, the computation of a posterior probability in a
Bayesian mixture model is intractable. The VB method provides an
analytical approximation solution for this problem
\cite{vsmidl2006variational,attias1999inferring}. In this section,
the proposed algorithms, the dSVB and dVB-ADMM, are applied to a
Bayesian Gaussian mixture model (GMM).

Consider a general wireless sensor network with $N$ nodes. Each
node $i$ has $N_i$ $D$-dimension measurements $\bm{x}_{ij}
(i=1,2,\dots,N, j=1,2,\dots,N_i)$. Due to the limitations on
energy and communication resources, we can not collect all the
data together, so we need to process the data distributedly. We
assume the measurements are modeled by a mixture of Gaussians with
$K$ components. Each component is a Gaussian distribution with the
mean $\bm{\mu}_k$ and covariance $\bm{\Lambda}_k^{-1}$,
\begin{equation*}
\mathcal{N}(\bm{x}_{ij}| \bm{\mu}_k, \bm{\Lambda}^{-1}_k)
= \frac{|\bm{\Lambda}_k|^{1/2}}{(2\pi)^{D/2}}
e^{-\frac{1}{2}(\bm{x}_{ij}-\bm{\mu}_k)^T \bm{\Lambda}_k (\bm{x}_{ij}-\bm{\mu}_k)}.
\end{equation*}
The Gaussian mixture distribution for observation $\bm{x}_{ij}$ is
\begin{equation} \label{std_gmm}
p(\bm{x}_{ij} | \bm{\pi}, \bm{\mu}, \bm{\Lambda}) =
\sum_{k=1}^K \pi_k \mathcal{N}(\bm{x}_{ij}| \bm{\mu}_k, \bm{\Lambda}^{-1}_k) ,
\end{equation}
where $\bm{\pi} = \{\pi_1,\dots,\pi_K\}$, $ \bm{\mu} = \{\bm{\mu}_1,\dots, \bm{\mu}_K\}$
and $\bm{\Lambda}=\{\bm{\Lambda}_1,\dots,\bm{\Lambda}_K\}$.
The standard mixture distribution (\ref{std_gmm}) does not belong to the exponential family
and therefore cannot be used directly as a conditional distribution
within a conjugate-exponential model.
Instead, we introduce an additional discrete latent variable $\{\bm{y}_i\}$ for each node,
which indicates from which component distribution each data point was drawn.
Hence, the local distribution at each node can be written as
\begin{equation}
P(\{\bm{x}_i\}_N|\bm{y}_i, \bm{\mu},\bm{\Lambda})
=\prod_{j=1}^{N_i} \prod_{k=1}^{K}
\mathcal{N}(\bm{x}_{ij}| \bm{\mu}_k, \bm{\Lambda}^{-1}_k)^{N \cdot y_{ijk}},
\end{equation}
where $\bm{y}_i = \{\bm{y}_{i1},\dots,\bm{y}_{iN_i}\}$ and $ \bm{y}_{ij} = \{y_{ij1},y_{ij2},\dots, y_{ijK}\}$.

The conjugate priors of the parameters and latent variables need to be specified.
The prior of $\bm{y}_i$ at node $i$ is a product of multinomials,
conditional on the mixing coefficients, which are assigned a Dirichlet prior.
The means are assigned multivariate Gaussian conjugate priors,
conditional on the precision matrices (inverse covariance matrices),
which are assigned Wishart priors. All priors are given by
\begin{subequations}
    \begin{align}
    P(\bm{y}_i|\bm{\pi}) &= \prod_{j=1}^{N_i}  \mbox{Mult}(1,\bm{\pi}),\\
    P(\bm{\pi}) &= \mbox{Dir}(K, \alpha_0), \label{mix_weight_prior} \\
    P(\bm{\mu}|\bm{\Lambda}) &= \prod_{k=1}^K \mathcal{N}(\bm{\mu}_0,(\beta_0 \bm{\Lambda}_k)^{-1}), \label{mean_prior}\\
    P(\bm{\Lambda}) &= \prod_{k=1}^K \mathcal{W}(\bm{W}_0,\nu_0). \label{prec_prior}
    \end{align}
\end{subequations}
With these priors, the generative model for each node $i$ is then factorized using the conditional independence,
\setlength{\arraycolsep}{0.0em}
\begin{eqnarray}
&&  P(\{\bm{x}_i\}_N, \bm{y}_i, \bm{\pi},\bm{\mu},\bm{\Lambda} ) \nonumber\\
&&= P(\{\bm{x}_i\}_N|\bm{y}_i, \bm{\mu},\Lambda) P(\bm{y}_i|\pi) P(\bm{\pi})P(\bm{\mu}|\bm{\Lambda}) P(\bm{\Lambda}).  \nonumber
\end{eqnarray}
\setlength{\arraycolsep}{5pt}
With the mean field assumption, the joint variational distribution of the unobserved variables is factorized as
\begin{equation*}
q(\bm{y}_i, \bm{\pi},\bm{\mu},\bm{\Lambda}) = q(\bm{y}_i) q(\bm{\pi}) \prod_{k=1}^K q(\bm{\mu}_k|\bm{\Lambda}_k)q(\bm{\Lambda}_k).
\end{equation*}
Using the VB update equation (\ref{opt_star}), the ``best''
variational distribution can be simply derived. As a general
result for a conjugate-exponential model, the form of each
distribution is the same as its prior. Specifically, the local
optimal variational distributions at node $i$ (only using local
data) are
\begin{subequations}
\begin{align}
    & q^*(\bm{y}_i) = \prod_{j=1}^{N_i} \mbox{Mult}(1,r_{ij1},\dots,r_{ijK}), \label{opt_mult}\\
    & q^*(\bm{\pi}_i) = \mbox{Dir}(\alpha_{i1},\dots,\alpha_{iK}), \label{opt_dir}\\
    & q^*(\bm{\mu}_{ik},\bm{\Lambda}_{ik})
     = \mathcal{N}(\bm{m}_{ik}, (\beta_{ik}\bm{\Lambda}_{ik})^{-1})
    \mathcal{W}(W_{ik},\nu_{ik}), \forall \, k, \label{opt_gas_wish}
\end{align}
\end{subequations}
where the update equations for the hyperparameters are given in
the Appendix \ref{apd_vbresult}. The parameters of
$q^*(\bm{y}_i)$ depend on the sufficient statistics of
$q^*(\bm{\pi}_i)$ and $q^*(\bm{\mu}_{ik},\bm{\Lambda}_{ik}),
\forall k$, whose parameters in turn depend on
the sufficient statistics of $q^*(\bm{y}_i)$.

To apply the dSVB and dVB-ADMM to a specific model, all we need to
do is to derive the local optimum of global natural parameters
(\ref{loc_opt}). Note that the natural parameters can be viewed as
a function of the hyperparameters, and the hyperparameters can be
simply transformed to a natural parameters. It's not necessary to
consider the latent variables $\{\bm{y}_i\}$ since they are local
variables. As for the mixing coefficients $\{\bm{\pi}_i\}$, the
natural parameter vector of the Dirichlet distribution is
\begin{equation*}
\bm{\phi}_{\bm{\pi}_i} = \left[\alpha_{i1}-1,\dots,\alpha_{iK}-1\right]^T.
\end{equation*}
The natural parameter vector of the normal-Wishart Distribution is given by
\setlength{\arraycolsep}{0.0em}
\begin{eqnarray}
\bm{\phi}_{\bm{\mu}_{ik},\bm{\Lambda}_{ik}} &=&
\left[
\begin{array}{c}
\frac{\nu_{ik}-D}{2}  \\
-\frac{1}{2}\bm{W}_{ik}^{-1} -\frac{\beta_{ik}}{2}\bm{m}_{ik}\bm{m}_{ik}^T \\
\beta_{ik}\bm{m}_{ik}  \\
-\frac{1}{2}\beta_{ik} \\
 \end{array}
\right], \forall k=1,\dots,K. \nonumber
\end{eqnarray}
\setlength{\arraycolsep}{5pt} To simplify the notation and to keep
notational consistency, we introduce a global natural parameter
vector $\bm{\phi}_{\bm{\theta},i}$ for the joint distribution
$q^*(\bm{\pi}_i)\prod_{k=1}^{K}q^*(\bm{\mu}_{ik},\bm{\Lambda}_{ik})$
(also in exponential family), defined as
\begin{equation} \label{full_natural_param}
\bm{\phi}_{\bm{\theta},i} = [ \bm{\phi}_{\bm{\pi}_i}^T,\bm{\phi}_{\bm{\mu}_{i1},
    \bm{\Lambda}_{i1}}^T,\dots,\bm{\phi}_{\bm{\mu}_{iK},\bm{\Lambda}_{iK}}^T ]^T.
\end{equation}
This global natural parameter vector  is the message to be
exchanged among nodes. As we see,  it is very different from those
in the literatures
\cite{safarinejadian2010distributed,weng2011diffusion,gu2008distributed},
who use the model parameters or sufficient statistics directly. In
our method, the natural parameter vector is properly scaled,
therefore we do not need to adjust them manually. Substituting the
update equations (in Appendix \ref{apd_vbresult}) of the
hyperparameters into (\ref{full_natural_param}), we can see that
these quantities are very close to the sufficient statistics, but
the prior information are included in our framework. In fact, the
natural parameter vector is a function of the expectation of
related sufficient statistics, as illustrated in Lemma
\ref{lem_opt_exp_fam}.

At the $t$th iteration of the distributed VB procedure at node
$i$, the local optimal hyperparameters of the global distributions are
denoted as $\{\alpha_{ik}^{*,t} \}$, $\{\bm{m}_{ik}^{*,t} \}$,
$\{\bm{\beta}_{ik}^{*,t} \} $, $\{\bm{W}_{ik}^{*,t} \}$ and
$\{\bm{\nu}_{ik}^{*,t} \}$ and summarized using a natural
parameter vector $\bm{\phi}_{\bm{\theta},i}^{*,t}$
(corresponding to the quantity in (\ref{loc_opt})). Once the
update equation of the natural parameter vector is derived and
substituted into Algorithm \ref{alg_dSVB} and Algorithm
\ref{alg_dVB_ADMM}, the dSVB and dVB-ADMM algorithms for
distributed inference/estimation of Gaussian mixtures are immediately
obtained, respectively.

\section{Experimental Results} \label{sec_exper_res}
In this section, the performance of the proposed VB algorithms for
distributed inference/estimation of Gaussian mixture model is
evaluated via numerical simulations on both synthetic and
real-world datasets.

\subsection{Performance of the distributed stochastic VB algorithm}
\begin{figure}[!t]
    \centering
    \includegraphics[width=2.6in]{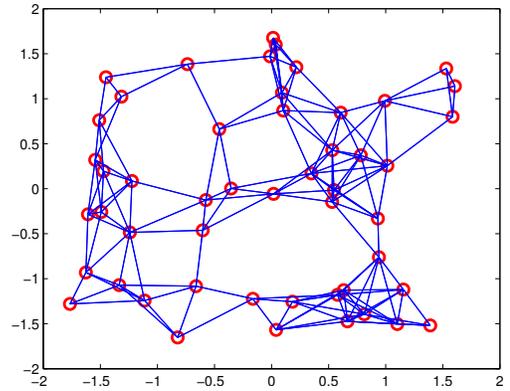}
    \caption{Network connection.}
    \label{fig_network}
\end{figure}

We consider a randomly generated sensor network with $50$ nodes.
The nodes are randomly placed in a  $3.5 \times 3.5$ square, and
the communication distance is taken as $0.8$. The constructed
connected network has $144$ edges, as shown in \figurename{\ref{fig_network}}. 
The 2-dimensional observations are
generated from the mixture of three Gaussian components ($K = 3$).
The corresponding parameter settings are as follows
\setlength{\arraycolsep}{0.0em}
\begin{eqnarray}
\bm{\pi} &{}={}& (0.32,0.45,0.23), \nonumber\\
\mu_1 &{}={}& (1.5,\ 3.5),\ \mu_2 = (4,\ 4),\ \mu_3 = (6.5,\ 4.5), \nonumber\\
\Sigma_1 &{}={}& \Sigma_3 =
\left[
\begin{array}{cc}
0.6,\ & 0.4 \\
0.4,\ & 0.6
\end{array}
\right],\
\Sigma_2  =
\left[
\begin{array}{cc}
0.6,\ & -0.4 \\
-0.4,\ & 0.6
\end{array}
\right]. \nonumber
\end{eqnarray}
\setlength{\arraycolsep}{5pt}

Each node has 100 data observations available ($N_i=100,
i=1,\dots,50$). In the first 15 nodes (node 1 to node 15), 80\%
observations come from the first Gaussian component and the other
20\% observations are evenly from the other two Gaussian components.
In the next 20 nodes (node 16 to node 35), 90\% observations come
from the second Gaussian component and the other 10\% are observations
evenly from the other two Gaussian components. In the last 15
nodes (node 36 to node 50), 60\% observations come from the third
Gaussian component and the other 40\% observations are evenly from the
other two Gaussian components.

The measure of performance used here is different from that used
in the previous works
\cite{safarinejadian2011distributed,forero2011distributed}.
Firstly, we point out that the local/global free energy can not
correctly assess the algorithm's performance, because the average
of all local lower bounds, which is obtained by maximizing each
local free energy independently, is always greater than or equal
to that obtained by maximizing the global free energy, as shown in
(\ref{global_less_local_bound}). Secondly, the simple mean squared
error (MSE) of estimates can not evaluate the algorithm's
performance well, since natural parameters differ greatly in
magnitude. The one that has the largest magnitude will impact the
MSE more than the others. The KL divergence is a good measure of
the difference between two distributions. It can measure the
information lost when using one distribution to approximate
another. In general, we can not use the KL divergence as the
measure of performance, since the true posterior is unknown.
However, in this {\it synthetic} example,
 we can compute the ground truth posterior of model parameters
 $P(\bm{\theta}|\hat{\bm{\phi}}_{\theta})$ in closed form
 based on the Bayes' theorem \cite{vsmidl2006variational},
 since the ground truth observation model belongs to exponential families and it has a conjugate prior.
Thus, we use the KL divergence between the joint variational
distribution of model parameters $
Q(\bm{\theta}|\bm{\phi}_{\theta,i}) $ estimated at each node and
the ground truth posterior
$P(\bm{\theta}|\hat{\bm{\phi}}_{\theta})$,
\begin{equation} \label{kl_cost}
d(\bm{\phi}_{\theta,i},\hat{\bm{\phi}}_{\theta})
= \mbox{KL}\big( Q(\bm{\theta}|\bm{\phi}_{\theta,i}) || P(\bm{\theta}|\hat{\bm{\phi}}_{\theta}) \big),
\end{equation}
to assess the local performance, and the mean of all KL
divergences to assess the global performance. Since $Q$ and $P$
are the same joint distribution (with different parameters)
belonging to the exponential families, the KL divergence can be
easily obtained in terms of a closed-form expression. The detailed
computation of the KL divergence is given in Appendix
\ref{apd_KL_div}.


\begin{figure}[!t]
    \centering
    \includegraphics[width=2.6in]{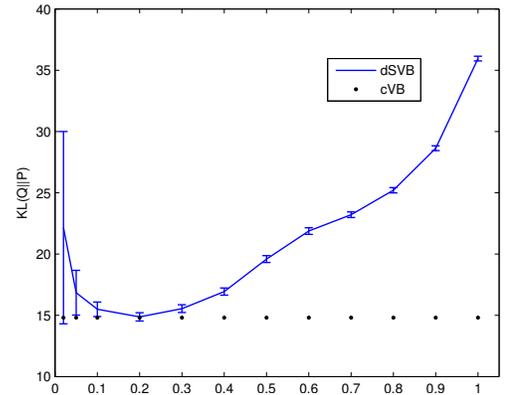}
    \caption{Error bars comparison between centralized VB and distributed SVB for various values of $\tau$.}
    \label{fig_dSVB_rho}
\end{figure}

For comparison, we simulate the centralized VB algorithm (cVB) for the GMM,
in which the global natural parameters (\ref{vbm_sol}) are computed by using all local quantities
in a fusion center. We also simulate the non-stochastic-gradient based distributed VB algorithm (nsg-dVB),
in which each node only uses its local optimum to diffuse the information with its neighbors.

There are two types of parameters needed to be determined for the
dSVB. The first is the combination weight. In all of the following experiments, we simply
assign it using the nearest neighbors rule \cite{xiao2004fast},
\begin{eqnarray}
w_{ij} \triangleq \left\{
    \begin{array}{ll}
    \frac{1}{|\mathcal{N}_i|+1},& \text{if} \ j \in \mathcal{N}_i \cup i \\
    0,&  \text{elsewise}
    \end{array}
    \right.
\end{eqnarray}
where $|\mathcal{N}_i|$ denotes the degree of node $i$.

The second is the forgetting rate $\tau$ for the step size
$\eta_t$ in (\ref{dSVB_step_size}). As analyzed in Section
\ref{subsec_dSVB}, the choice of  the forgetting rate $\tau$ is
important for the performance of the dSVB. We next show how the
performance depends on the value of $\tau$ numerically.
\figurename{\ref{fig_dSVB_rho}} shows the means and the standard
deviations of the cost (\ref{kl_cost}) among all $50$ nodes
obtained after $t=2000$ iterations for different values of the
forgetting rate $\tau$ with the same initialization. For
comparison, \figurename{\ref{fig_dSVB_rho}} also shows the cost of
the centralized VB. As we see, the mean of the cost is
approximately minimized in the interval $[0.1,0.3]$ and making it
either smaller or larger will lead to a higher cost. The standard
deviation of the cost measures the difference of estimates among
all nodes. This test reveals that the standard deviation goes down
when the forgetting rate increases, because the step size $\eta_t$
becomes smaller, so as the residuals in the second term of
(\ref{dSVB_single_update}). Taking both the mean and standard
deviation of the cost into account, we choose $\tau=0.2$ in the
following simulations.

\begin{figure}[!t]
    \centering
    \includegraphics[width=2.6in]{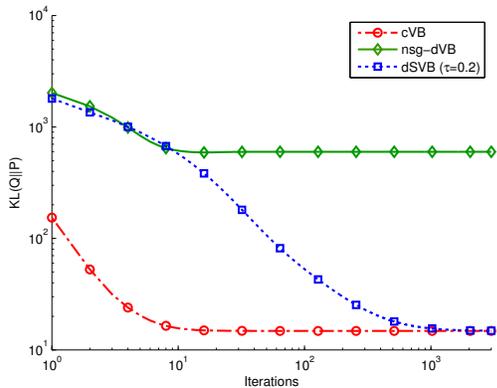}
    \caption{The evolution of the mean cost (KL divergence) among all nodes
    of the dSVB($\tau=0.2$), compared with the cVB and nsg-dVB. }
    \label{fig_cmp_dSVB_evolution}
\end{figure}

In the second simulation, the performance of the dSVB is tested
and compared with the cVB and nsg-dVB.
\figurename{\ref{fig_cmp_dSVB_evolution}} shows the evolution of
the mean cost (KL divergence) with the iterations. 
The non-stochastic-gradient based VB (nsg-dVB)
algorithm gets stuck at a local optimum and induces a very large
bias, because the previous combination effect is eliminated by the
VBE step and only the local information is utilized in every VBM
step. Unlike the nsg-dVB, the dSVB improves the estimates
gradually with the information diffused over the entire network.
Finally the dSVB reaches a result as good as the centralized VB.

\begin{figure*}[htb]
    \centering
    \includegraphics[width=6.5in]{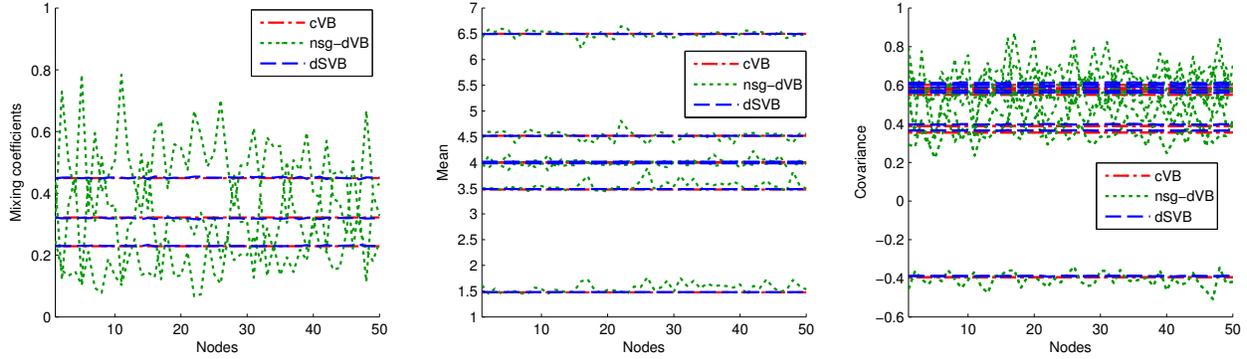}
    \caption{The values of the estimated mixing coefficients (left), means (middle) and covariances (right)
        using the cVB, nsg-dVB and dSVB.
        The vector (means) and matrices (covariance) values are visualized
        by their entries for a better comparison in 2D coordinates.}
    \label{fig_cmp_dSVB_value}
\end{figure*}

\begin{figure}[!t]
    \centering
        \includegraphics[width=2.6in]{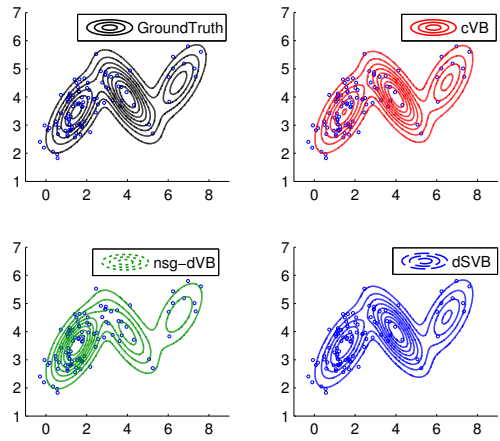}
    \caption{The contours of the estimated mixture model at a
        randomly selected node (node 67) using the dSVB and nsg-dVB
        compared with the cVB's.}
    \label{fig_cmp_dSVB_contour}
\end{figure}

In \figurename{\ref{fig_cmp_dSVB_value}}, the final estimates
among all nodes obtained after $3000$ iterations (the dSVB
 converges after $1000$ iterations in most cases) are
compared with those obtained by the cVB and nsg-dSVB. It can be seen that the
estimates obtained by the nsg-dVB are very different among all
nodes. In contrast, the estimate obtained by the dSVB at each node
is very close to that obtained by the centralized VB. It is worth
pointing out that although the local observed data at each node is
imbalanced, all local imbalanced data together is balanced. The
dSVB scheme can maintain a balance between the information from
local data and that from neighbors. As an example,
\figurename{\ref{fig_cmp_dSVB_contour}} shows the contours of
models estimated by different approaches at a randomly selected
node. As we see, the nsg-dVB can not correctly estimate the
mixture model, and it is still strongly impacted by the local
imbalanced data. As for the dSVB, the estimated model is almost
the same as the cVB's and the ground truth. The other nodes have
similar results, which are not given due to space limitations.

\subsection{Performance of the distributed VB-ADMM algorithm}
In this subsection, the performance of the dVB-ADMM
is tested using the same configuration established in the previous subsection.

\begin{figure}[!t]
    \centering
    \includegraphics[width=2.6in]{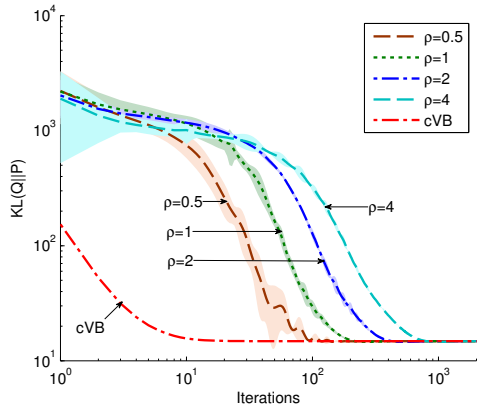}
    \caption{The evolution of the means and standard deviations of the cost
            (KL divergence) of the dVB-ADMM algorithm
             with different penalty parameters $\rho$, compared with the cVB.}
    \label{fig_cmp_dVB_admm_rho}
\end{figure}

The first experiment explores the convergence property of the
dVB-ADMM with different values of the penalty parameter $\rho$.
\figurename{\ref{fig_cmp_dVB_admm_rho}} reveals that a small value
of $\rho$ can give faster convergence. While larger values of
$\rho$ ensure that the natural parameters per node achieve the
same value faster, since it produces smaller primal residuals
$\{\bm{r}_i^t\}$, as we have analyzed in \textit{Remark}
\ref{remark_rho}. Note that if the value of $\rho$ is too small,
the primal residuals among nodes may become very large at the very
beginning stage of VB iterations, which may give more chance to
natural parameters to be out of the domain $\Omega$. In this
simulation, the covariance matrices of Gaussian components
sometimes become negative definite when $\rho<0.5$ (without
projection). Therefore, the value of $\rho$ should be small for
the fast convergence speed but not too small. Unless otherwise
specified, we choose $\rho=0.5$ in the following experiments.

\begin{figure}[!t]
    \centering
    \includegraphics[width=2.6in]{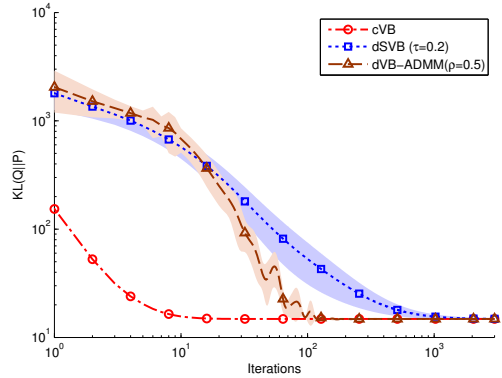}
    \caption{The evolution of the mean and standard deviation of the cost
        (KL divergence) of the dVB-ADMM ($\rho=0.5$), compared with the dSVB ($\tau=0.2$) and the cVB. }
    \label{fig_dVBADMM_dSVB}
\end{figure}

Next, we compare the performance of the dVB-ADMM with those of the
cVB and dSVB. \figurename{\ref{fig_dVBADMM_dSVB}} shows
that the dVB-ADMM outperforms the dSVB both in the convergence
speed and the accuracy. Using the dSVB, the cost and differences
among nodes decrease gradually and smoothly. While, in the
dVB-ADMM, the estimate values and their differences
among nodes fluctuate within a wide range at the beginning stage of
iterations, but become very stable after a few hundreds of
iterations. The dSVB gets converged after about $1000$ iterations,
while the dVB-ADMM only needs about $200$ iterations to reach the
same accuracy.

\subsection{Evaluation of the robustness of the algorithms}
In the above experiments, the distribution that draws the local
sampled data from is very different for different nodes, but the
number of the local sampled data points at each node is assumed 
to be the same. In this setting, both the dSVB and the dVB-ADMM approaches
perform well. However, in many practical cases, the number of the
sampled data point at each node could be very different. In order
to test the robustness of the proposed algorithms, we evaluate
their performance in this case. Furthermore, note that the number
of nodes in different networks may be very different either, we will
show that our approaches are scalable and can be applied into networks with
different sizes.

\begin{figure}[!t]
    \centering
    \includegraphics[width=2.6in]{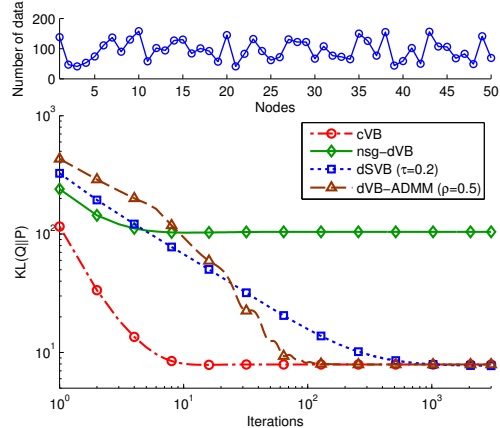}
    \caption{The number of data at each node (top) and the average performance
        (bottom) of the dSVB and dVB-ADMM with the imbalanced data, compared
        with the nsg-dVB and cVB. }
    \label{fig_UnequalDataNum}
\end{figure}

\subsubsection{The case of the unequal data sizes} In this experiment,
we consider the case that the observed data sizes among different
nodes are unequal. They are randomly selected from $40\sim 160$,
as shown in the top panel of
\figurename{\ref{fig_UnequalDataNum}}. The distribution parameters
are kept the same as those in the above simulations. All data
samples are randomly generated from the whole Gaussian mixture
model. In this setting, \figurename{\ref{fig_UnequalDataNum}}
illustrates that the dSVB and dVB-ADMM can also perform much
better than the nsg-dVB and almost as well as the centralized VB.
The unbalancedness of the sample size has no significant impact on
the performance. In fact, in the process of deriving the
algorithms, we have never made any assumption about the number of
the local data, and the natural parameter vector already carries
this information. This again explains why the natural parameter
vector is a good choice of the message exchanged among nodes.

\subsubsection{The case of different network sizes}
\begin{figure}[!t]
    \centering
    \includegraphics[width=2.6in]{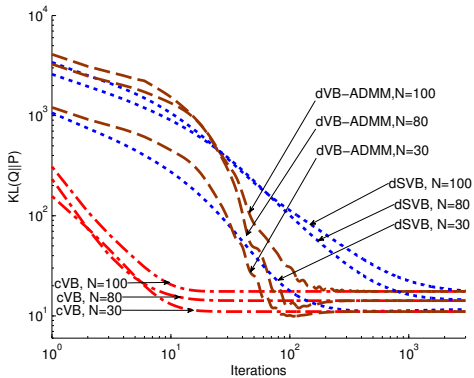}
    \caption{The average performance of the dSVB and dSV-ADMM with different
        network sizes ($N=30,80,100$). The tuning parameters are set as $\tau=0.2$, $\rho=0.5$ for all cases.}
    \label{fig_varynodenum}
\end{figure}

In the situation with different network sizes, the performance of
both algorithms are evaluated here. The density of networks
remains unchanged. To ensure this, the communication distance
still remains $0.8$ and the square, where the nodes are randomly
placed, is proportionally zoomed in and out. Other settings are
kept unchanged. We test various sizes of networks and three of
them ($N=30,80,100$) are shown in \figurename{\ref{fig_varynodenum}}. 
With the increase of the network size, the total number of iterations needed to get
converged increases. It's inevitable because the local
data is partial and the full information is distributed to all
nodes. Nevertheless, the convergence can still be achieved and the
performance is still good.

\subsection{Clustering of real data}
In this subsection, we examine the proposed algorithms for
distributed clustering using GMM on three real-world datasets. We
use the accuracy or the misclassification rate to measure the
clustering performance. For comparison, except simulating the cVB
and nsg-VB, we also simulate the non-cooperation VB algorithm
(noncoop-VB), in which each node performs the VB without
cooperating with its neighbors.

\subsubsection{Atmosphere Data}
\begin{figure}[!t]
    \centering
    \includegraphics[width=2.6in]{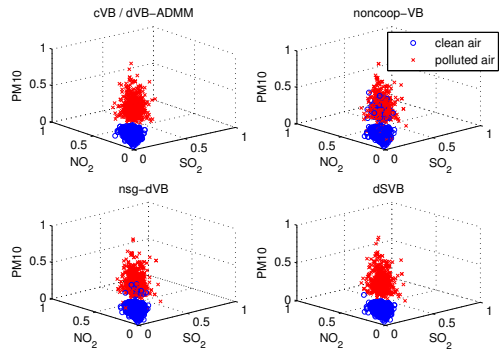}
    \caption{Clustering results of different algorithms on the atmosphere dataset in one trial.}
    \label{fig_atmos}
\end{figure}

To get an overall evaluation of the atmosphere quality, we can
collect air samples distributedly using a WSN. The evaluation task
is then taken by utilizing local computation and one-hop
communication in the WSN. In this experiment, we use the real
atmosphere data provided by \cite{shen2014distributed}. A total of
$1600$ samples (including $830$ clean air and $770$ polluted air
samples) are used, each with entries, the sulfur dioxide ($SO_2$),
nitrogen dioxide ($NO_2$) and PM10. We use a WSN with $20$ nodes,
each with randomly allocated $N=80$ measurements. The algebraic
connectivity of the network is $0.24$ and the average degree is
$4.8$. For this clustering task, the tuning parameters for the two
proposed algorithms are set as $\tau = 0.2$ and $\rho = 1$,
respectively. Table \ref{tab_atmos} shows the average accuracy and
the number of misclassification samples for different algorithms over
300 independent Monte Carlo simulations with random
initializations. \figurename{\ref{fig_atmos}} depicts the
clustering results of different algorithms in one trial. From
these results, we see that the numbers of the misclassification samples 
of the dSVB and the dVB-ADMM are much lower than that of the noncoop-VB and
the nsg-dVB while very close to that of the cVB.

\begin{table}[!t]
    \renewcommand{\arraystretch}{1.3}
    \caption{The accuracy and average number of misclassification samples of different algorithms on the atmosphere data}
    \label{tab_atmos}
    \centering
    \begin{tabular}{|c||c|c|}
        \hline
        Algorithm & Accuracy (\%) & \# misclassification samples \\  \hline \hline
        cVB        & $\mathbf{100.00} $ & $\mathbf{0.00}$ \\ \hline
        noncoop-VB & $89.75$ & $164.06$ \\ \hline
        nsg-dVB    & $98.99$ & $16.15$ \\ \hline
        dSVB       & $99.89$ & $1.73$ \\ \hline
        dVB-ADMM   & $99.99$ & $0.03$ \\ \hline
    \end{tabular}
\end{table}

\subsubsection{Ionosphere Data}
To give an overall analysis of the ionosphere, a radar sensor
network can be used to collect the radar returns from the
ionosphere and cooperatively figure out which ones are ``good''
(showing evidence of some type of structure in the ionosphere) and
which are ``bad''. To simulate this scenario, we perform the
algorithms on the ionosphere dataset from the UCI learning
repository \cite{sigillito1989classification}. This radar data was
collected from a phased array of sixteen high-frequency antennas
in Goose Bay, Labrador. There are $351$ observations (including
$225$ ``good'' and $126$ ``bad'' radar returns) with $34$
continuous attributes. We use the same sensor network as that in
the previous experiment. The algorithms' parameters are set as
$\tau = 0.2$ and $\rho = 16$. The evaluation is performed by
averaging over $300$ independent Monte Carlo simulations with
random initializations. For each simulation, $340$ observations
are randomly selected from the whole dataset and uniformly
distributed to $20$ nodes. As shown in Table \ref{tab_ionos}, the
dSVB and the dVB-ADMM outperform the noncoop-VB and the nsg-dVB.
Surprisingly, the dVB-ADMM can even get higher accuracy than the
centralized VB.

\begin{table}[!t]
    \renewcommand{\arraystretch}{1.3}
    \caption{The accuracy and average number of misclassification samples of different algorithms on the ionosphere data}
    \label{tab_ionos}
    \centering
    \begin{tabular}{|c||c|c|}
        \hline
        Algorithm & Accuracy (\%) & \# misclassification samples \\  \hline \hline
        cVB        & $82.08$ & $60.94$ \\ \hline
        noncoop-VB & $59.65$ & $137.19$ \\ \hline
        nsg-dVB    & $64.89$ & $119.36$ \\ \hline
        dSVB       & $78.25$ & $73.96$ \\ \hline
        dVB-ADMM   & $\mathbf{85.59}$ & $\mathbf{49.00}$ \\ \hline
    \end{tabular}
\end{table}

\subsubsection{COIL-20 Data} 
\begin{figure}[!t]
    \centering
    \includegraphics[width=2.6in]{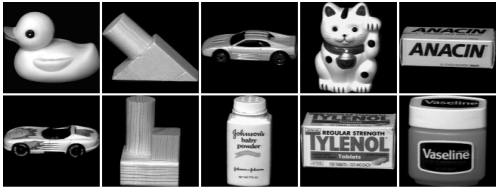}
    \caption{Sample images from the COIL-20 dataset.}
    \label{fig_coil_sample}
\end{figure}

In many practical image processing applications,
in order to obtain multi-aspect rich information, images are often
acquired from many different positions in an environment. In this
case, a WSN with sensor nodes equipped with tiny cameras can be
used. To simulate the above scenario, we use the proposed
algorithms for cooperatively solving an object classification
problem using the COIL-20 image dataset \cite{nene1996columbia}.
It contains $20$ objects. The images of each objects were taken
$5$ degrees apart as the object is rotated on a turntable and each
object has $72$ images. Some sample images are shown in
\figurename{\ref{fig_coil_sample}}. The size of each image is $32
\times 32$ pixels, with $256$ gray levels per pixel. Thus, each
image is represented by a 1,024-dimensional vector. To speed up
iteration and to avoid the covariance matrix being singular, we
apply PCA to reduce the dimension to $52$, which keeps about $90$
percent information according to the eigenvalues. We use a WSN
with $10$ nodes. The algebraic connectivity is $0.51$ and the
average degree per node is $3.0$. For each given cluster number
$K$ (ranges from $2$ to $10$), $30$ tests are conduced on randomly
chosen clusters. For each test, the corresponding images are
randomly grouped into $10$ subsets and uniformly allocated to the
$10$ nodes. The tuning parameters are set as $\tau = 0.2$, $\rho =
16$.

\begin{figure}[!t]
    \centering
    \includegraphics[width=2.6in]{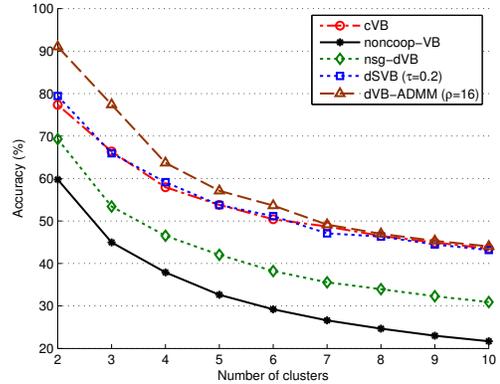}
    \caption{Accuracy versus the number of clusters on the COIL-20 dataset.}
    \label{fig_coil_acc}
\end{figure}

The final performance scores for each clustering number $K$ were computed by averaging the scores
from $300$ independent tests with random initializations.
\figurename{\ref{fig_coil_acc}} shows the plots of the clustering performance versus the number of clusters.
As we see, our proposed dSVB and dVB-ADMM algorithms perform
almost as well as the centralized VB (cVB), and much better than
the nsg-dVB and noncoop-VB. In some cases (when $K<7$), the
dVB-ADMM can achieve better clustering results than the
centralized VB. The reason is that our distributed algorithms can
be seen as the sparse and incremental variants of the VB algorithm
\cite{neal1998view}, which might be less sensitive to
initialization and on average they can find better local optimum
more often than their centralized counterparts with random
initializations. Thus, distributed algorithm with multiple parts
of data might have more advantage to avoid obtaining a worse local
optimum than centralized algorithm with the whole data. Similar
phenomena have also been observed in other distributed clustering
algorithms \cite{shen2014distributed,forero2011distributed}.

\section{Conclusion} \label{sec_conclusions}
In this paper, two distributed variational algorithms are proposed
for general Bayesian inference in a networked system. The
variational problem is recast as an optimization problem in the
natural parameter space. The variational free energy is decomposed
as a set of local lower bounds. Each node runs the VBE step by
maximizing the local lower bound with respect to the local latent
variables. The VBM step is solved in two distributed schemes. In
the dSVB scheme, a stochastic natural gradient is adopted for
gradually improving the estimates with the local data and a
combination step is used for the cooperation with neighbors. In
the dVB-ADMM scheme, the variational optimization is redefined as
a constrained minimization problem with a modified objective
function. The ADMM technique is then used to solve this
constrained optimization. In both schemes, each node only needs to
exchange low dimensional natural parameters with its neighbors.

An application of the distributed inference/estimation of a
Bayesian Gaussian mixture model is then presented, to evaluate the
effectiveness of the proposed algorithms. The numerical
simulations on both synthetic and real-world datasets demonstrate
that the proposed algorithms outperform the
non-stochastic-gradient based distributed VB (nsg-dVB) and
non-cooperation VB (noncoop-VB) algorithms and they both can
perform almost as well as the centralized VB (cVB). Both of the
algorithms exhibit resilience to data imbalance and is
applicable to networks with different sizes. The dVB-ADMM approach
converges faster than the dSVB, but even the dSVB algorithm
performs much better than the competing algorithms in the
literature.

Further development includes expanding results to the
non-conjugate exponential family and developing distributed
algorithms for dynamic Bayesian networks.


%

\appendix

\subsection{Distributed VB for Gaussian mixtures} \label{apd_vbresult}
The hyperparameters for (\ref{opt_mult}), (\ref{opt_dir}) and (\ref{opt_gas_wish}) are
\begin{equation*}
    \begin{split}
    & r_{ijk} =  \rho_{ijk} / \sum_{k=1}^K  \rho_{ijk}, \bm{m}_{ik} 
    = \frac{1}{\beta_{ik}} (\beta_0 \bm{\mu}_0 + R_{ik} \bar{\bm{x}}_{ik} ), \\
    &\alpha_{ik} = \alpha_0 + R_{ik},\ \beta_{ik}
    = \beta_0 + R_{ik},\ \nu_{ik}  = \nu_0 + R_{ik},  \\
    &\bm{W}_{ik}^{-1} = \bm{W}_0^{-1} + R_{ik} \bm{S}_{ik} +
    \frac{\beta_0 R_{ik}}{\beta_0 + R_{ik}}(\bar{\bm{x}}_{ik}-\bm{\mu}_0)(\bar{\bm{x}}_{ik} -\bm{\mu}_0) ^T,
    \end{split}
\end{equation*}
where the sufficient statistics are given by
\begin{equation*}
\begin{split}
\ln \rho_{ijk} & = \mathbb{E}[\ln {\pi}_{ik}]
+ \frac{1}{2} \mathbb{E}[\ln |\bm{\Lambda}_{ik}|]
- \frac{D}{2} \ln (2 \pi) \nonumber\\
&- \frac{1}{2} \ln \mathbb{E}_{\bm{\mu}_{ik},\bm{\Lambda}_{ik}} [(\bm{x}_{ij}
- \bm{\mu}_{ik})^T \bm{\Lambda}_{ik}(\bm{x}_{ij} - \bm{\mu}_{ik})],\\
R_{ik} &= N \sum_{j=1}^{N_i} r_{ijk}, \
\bar{\bm{x}}_{ik}  = \frac{N}{R_{ik}} \sum_{j=1}^{N_i} r_{ijk} \bm{x}_{ij}, \\
\bm{S}_{ik} &=\frac{N}{R_{ik}} \sum_{j=1}^{N_i} r_{ijk} (\bm{x}_{ij} -  \bar{\bm{x}}_{ik}) (\bm{x}_{ij} -  \bar{\bm{x}}_{ik})^T.
\end{split}
\end{equation*}
The expectations in the above
formulae are derived using (\ref{first_order_stat}),
\begin{equation*}
    \begin{split}
    &\mathbb{E}[\ln {\pi}_{ik}]  =  \psi(\alpha_{ik}) - \psi\left(\sum_{i=1}^{K} \alpha_{ik}\right), \nonumber\\
    & \mathbb{E}[\ln |\bm{\Lambda}_{ik}|] =\sum_{j=1}^{D}
    \psi\left(\frac{\nu+1-j}{2}\right) + D \ln 2 + \ln |\bm{W}_{ik}|, \nonumber\\
    &\mathbb{E}_{\bm{\mu}_{ik},\bm{\Lambda}_{ik}}
    [(\bm{x}_{ij} - \bm{\mu}_{ik})^T \bm{\Lambda}_{ik}(\bm{x}_{ij} - \bm{\mu}_{ik})]  =  D\beta_{ik}^{-1}  \nonumber\\
    & {\ \ \ \ \ \ \ \ \ \ \ \ \ \ \ \ \ \ \ \ \ \ \ \ }+
    \nu_{ik}(\bm{x}_{ij}-\bm{m}_{ik})^T \bm{W}_{ik}(\bm{x}_{ij}-\bm{m}_{ik}), \nonumber
    \end{split}
\end{equation*}
where $\psi(\cdot)$ is digamma function.

\subsection{KL divergence for exponential family distributions} \label{apd_KL_div}
We omit the subscript $i$, the node index, for notational
simplicity. The joint variational distribution of model parameters
in the synthetic example in Section V-A is
\begin{equation*} 
Q(\bm{\pi},\bm{\mu},\bm{\Lambda}|\bm{\phi}_{\theta}) = q(\bm{\pi})\prod_{k=1}^K q(\bm{\mu}_k,\bm{\Lambda}_k), 
\end{equation*}
and the corresponding ground truth posterior is
\begin{equation*}
P(\bm{\pi},\bm{\mu},\bm{\Lambda}|\hat{\bm{\phi}_{\theta}}) = P(\bm{\pi})\prod_{k=1}^K P(\bm{\mu}_k,\bm{\Lambda}_k),
\end{equation*}
where $q(\bm{\pi}) = \mbox{Dir}(\bm{\pi}|\bm{\alpha})$, $P(\bm{\pi}) =  \mbox{Dir}(\bm{\pi}|\hat{\bm{\alpha}})$ are Dirichlet distributions,
and $q(\bm{\mu}_k,\bm{\Lambda}_k) = \mathcal{NW}(\bm{\mu}_k,\bm{\Lambda}_k |\bm{\phi}_k)$,
$P(\bm{\mu}_k,\bm{\Lambda}_k) = \mathcal{NW}(\bm{\mu}_k,\bm{\Lambda}_k |\hat{\bm{\phi}}_k)$ are normal-Wishart distributions.
Hence, the KL divergence between two distributions becomes
\begin{equation*}
\begin{split}
& d(\bm{\phi}_{\theta},\hat{\bm{\phi}}_{\theta}) \\
&= \mbox{KL}\big(Q(\bm{\pi},\bm{\mu},\bm{\Lambda}|\bm{\phi}_{\theta}) || P(\bm{\pi},\bm{\mu},\bm{\Lambda} |\hat{\bm{\phi}}_{\theta}) \big) \\
& = \mbox{KL}(\mbox{Dir}(\bm{\pi}|\bm{\alpha}) || \mbox{Dir}(\bm{\pi}|\hat{\bm{\alpha}})) \nonumber\\
& \quad + \sum_{k=1}^{K}\mbox{KL}(\mathcal{NW}(\bm{\mu}_k,\bm{\Lambda}_k|\bm{\phi}_k)
|| \mathcal{NW}(\bm{\mu}_k,\bm{\Lambda}_k |\hat{\bm{\phi}}_k).
\end{split}
\end{equation*}
The KL divergences in the above equation can be computed as
follows.
\subsubsection{Dirichlet distribution}
The KL divergence between two Dirichlet distributions is
\begin{equation}
\begin{split}
&\mbox{KL}(\mbox{Dir}(\bm{\pi}|\bm{\alpha}) || \mbox{Dir}(\bm{\pi}|\hat{\bm{\alpha}})) \nonumber\\
& = \sum_{k=1}^K(\bm{\alpha}_{k}-\hat{\bm{\alpha}}_{k}) \mathbb{E}_{\bm{\alpha}}[\ln \bm{\pi}_{k}] -
\ln {B}(\bm{\alpha}) + \ln {B}(\hat{\bm{\alpha}}),
\end{split}
\end{equation}
where $\bm{\alpha} = [\bm{\alpha}_1,\dots,\bm{\alpha}_K]^T$, $\bm{\pi} = [\bm{\pi}_1,\dots,\bm{\pi}_K]^T$ and ${B}(\cdot)$ is the multinomial Beta function,
which can be expressed as ${B}(\bm{\alpha}) = \prod_{k=1}^{K}\Gamma(\bm{\alpha}_k)/ \Gamma(\sum_{k=1}^K \bm{\alpha}_k)$.

\subsubsection{Normal-Wishart distribution}
The KL divergence between two normal-Wishart distributions is
\begin{equation}
\begin{split}
&\mbox{KL}(\mathcal{NW}(\bm{\mu}_k,\bm{\Lambda}_k|\bm{\phi}_k)
|| \mathcal{NW}(\bm{\mu}_k,\bm{\Lambda}_k |\hat{\bm{\phi}}_k) \nonumber\\
& =  \operatorname{tr} \left( [\bm{\phi}_k - \hat{\bm{\phi}}_k]^T  \left[
\begin{array}{c}
\mathbb{E}_{\bm{\phi}_k}[\ln |\bm{\Lambda}_k|] \\
\mathbb{E}_{\bm{\phi}_k}[\bm{\Lambda}_k] \\
\mathbb{E}_{\bm{\phi}_k}[\bm{\Lambda}_k \bm{\mu}_k]\\
\mathbb{E}_{\bm{\phi}_k}[\bm{\mu}_k^T\bm{\Lambda}_k \bm{\mu}_k]
\end{array}
\right] \right) - A(\bm{\phi}_k) + A(\hat{\bm{\phi}}_k), \nonumber
\end{split}
\end{equation}
where the natural parameter vector for normal-Wishart distribution
is
$$\bm{\phi}_k = [\frac{\nu_k-D}{2},
-\frac{1}{2}\bm{W}_k^{-1}-\frac{\beta_k}{2}\bm{m}_k\bm{m}_k^T,
\beta_k\bm{m}_k, -\frac{1}{2}\beta_k]^T,$$ and the sufficient
statistics that have not been given in Appendix \ref{apd_vbresult}
include
\begin{subequations}
    \begin{align}
    \mathbb{E}_{\bm{\phi}_k}[\bm{\Lambda}_k] &= \nu_k \bm{W}_k, \nonumber\\
    \mathbb{E}_{\bm{\phi}_k}[\bm{\Lambda}_k \bm{\mu}_k] &=  \nu_k \bm{W}_k \bm{m}_k, \nonumber\\
    \mathbb{E}_{\bm{\phi}_k}[\bm{\mu}_k^T \bm{\Lambda}_k \bm{\mu}_k] &= D \beta_k^{-1} + \nu_k\bm{m}_k^T \bm{W}_k \bm{m}_k. \nonumber
    \end{align}
\end{subequations}
In addition, the partition function $A(\cdot)$ for this
distribution is
\begin{equation}
\begin{split}
A(\bm{\phi}_k) & = -\frac{D}{2}\ln |\beta_k|
+ \frac{\nu_k}{2} \ln |\bm{W}_k| +\frac{\nu_k D}{2} \ln 2 \nonumber\\
& + \sum_{j=1}^D \ln \Gamma(\frac{\nu_k+1-j}{2}). \nonumber
\end{split}
\end{equation}
Hence, the KL divergence between $P$ and $Q$ can be computed
in terms of a closed-form expression using the above equations.


\ifCLASSOPTIONcaptionsoff
  \newpage
\fi




%

\begin{IEEEbiography}	[{\includegraphics[width=1in,height=1.25in,clip,keepaspectratio]{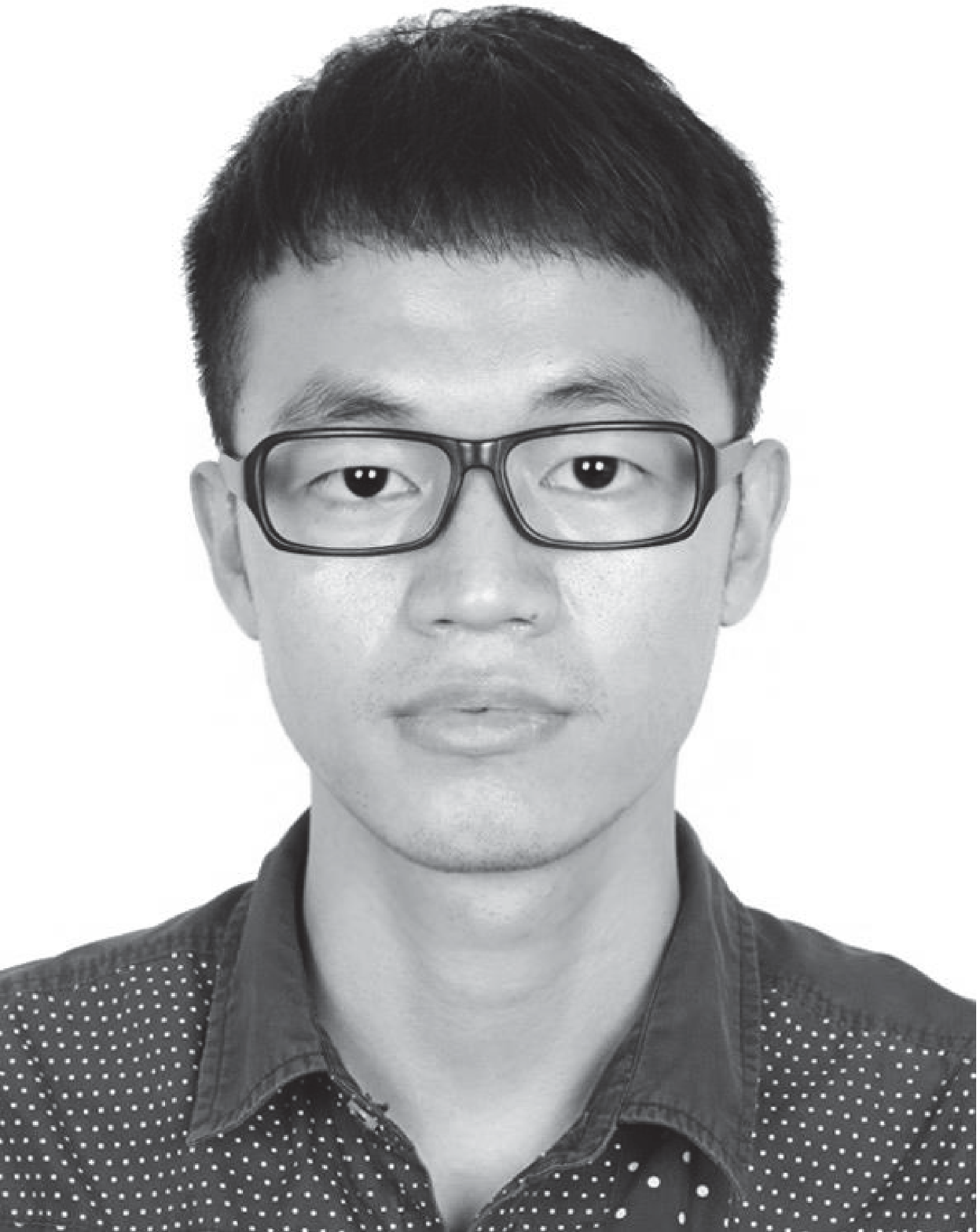}}]{Junhao Hua}
	received the B.S. degree both in computer
	science and  automation from Zhejiang
	University of Technology, Hangzhou, China, in 2013.
	
	Currently, he is pursuing the Ph.D. degree in the
	College of Information Science and Electronic
	Engineering, Zhejiang University, Hangzhou, China.
	His current research interests include statistical signal
	processing, Bayesian learning and wireless sensor network.
\end{IEEEbiography}

\begin{IEEEbiography}
	[{\includegraphics[width=1in,height=1.25in,clip,keepaspectratio]{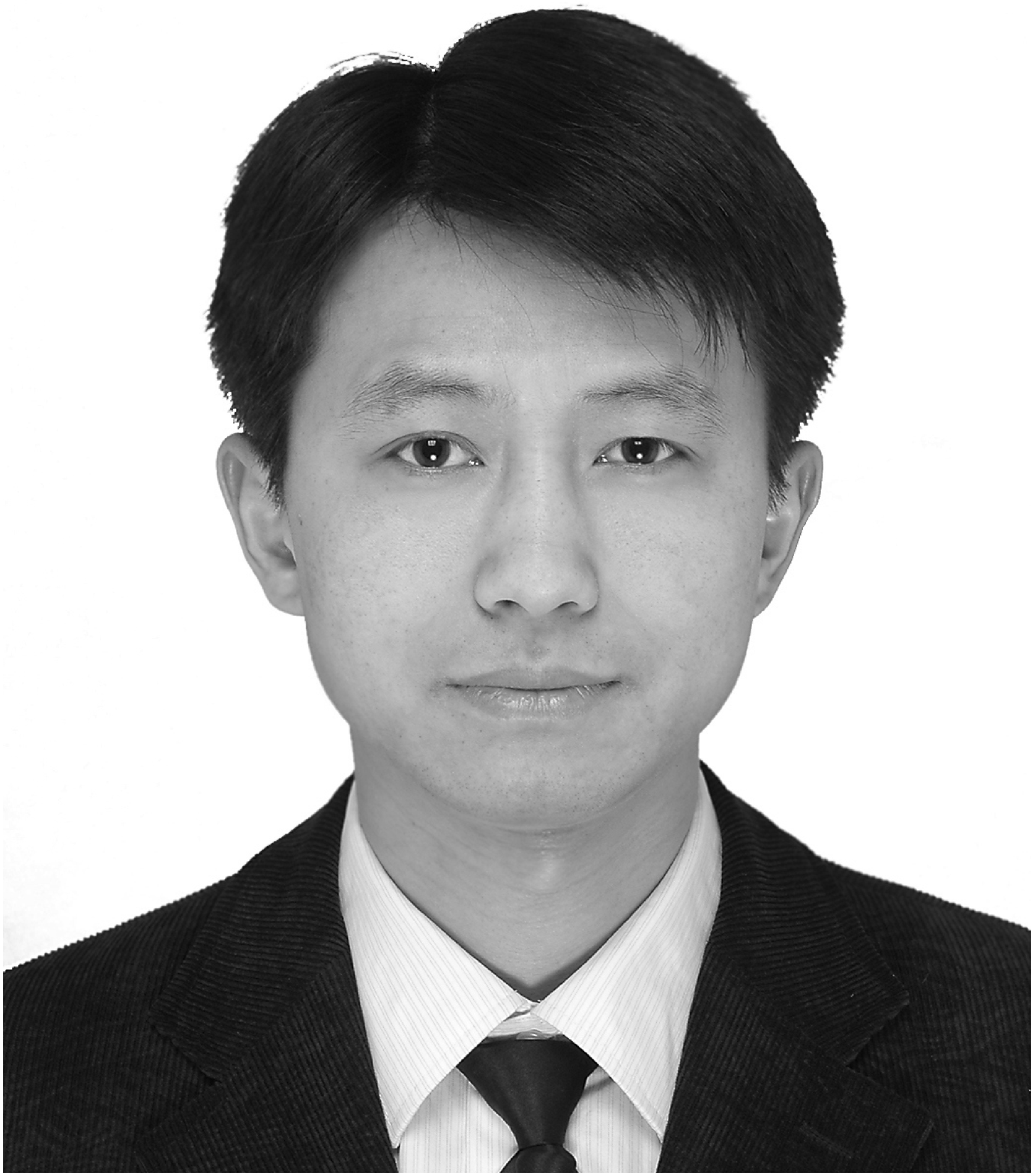}}]{Chunguang Li}(M'14--SM'14)
	received the M.S. degree in Pattern Recognition and Intelligent Systems and the Ph.D. degree 
	in Circuits and Systems from the University of Electronic Science and Technology of China, 
	Chengdu, China, in 2002 and 2004, respectively.
	
	Currently, he is a Professor with the College of Information Science and Electronic Engineering, 
	Zhejiang University, Hangzhou, China. His current research interests include statistical 
	signal processing and wireless sensor network.
\end{IEEEbiography}

\end{document}